\documentclass[preprint,12pt,3p]{elsarticle}

\usepackage{hyperref}
    
\usepackage{times}
\usepackage{amssymb, amsmath, amsbsy}
\usepackage{makecell}  
\usepackage{graphicx}
\usepackage{float}
\usepackage{multirow}
\usepackage{xcolor}
\usepackage{soul}
\usepackage[capitalise]{cleveref}
\crefformat{appendix}{#2#1#3}

\usepackage{natbib}
\newcommand{\bs}[1]{\boldsymbol{#1}}
\newcommand{\mc}[1]{\mathcal{#1}}
\newcommand{\bmc}[1]{\boldsymbol{\mathcal{#1}}}

\newcommand{\grad}[0]{\bs{\nabla}}
\newcommand{\gradd}[0]{\bar{\bs{\nabla}}}
\newcommand{\divg}[0]{\bs{\nabla}\cdot}
\newcommand{\divgd}[0]{\bar{\bs{\nabla}}\cdot}
\newcommand{\partt}[1]{\frac{\partial #1}{\partial t}}
\newcommand{\parttd}[1]{\frac{\partial #1}{\partial \bar{t}}}

\newcommand{\dyad}[0]{\otimes}
\newcommand{\tr}[0]{\mathrm{tr}}

\usepackage{amsmath}

\DeclareMathOperator*{\argmin}{arg\,min}

\usepackage{algorithm}
\usepackage{algpseudocode}

\journal{Elsevier}

\begin{document}

\begin{frontmatter}

\title{Physics-informed neural network simulation of multiphase poroelasticity using stress-split sequential training}

\author[MIT,UBC]{
\href{https://orcid.org/0000-0003-2659-0507}{\includegraphics[scale=0.06]{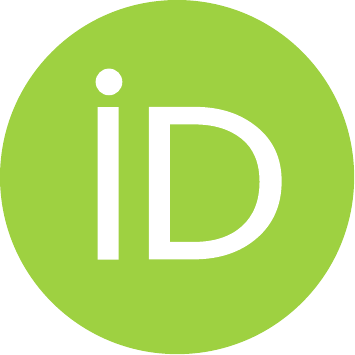}\hspace{1mm}Ehsan Haghighat}\corref{cor1}}
\address[MIT]{Department of Civil and Environmental Engineering, Massachusetts Institute of Technology, Cambridge MA, US}
\address[UBC]{Department of Civil Engineering, University of British Columbia, Vancouver BC, Canada}
% \ead{ehsanh@mit.edu}
% \ead[url]{author-one-homepage.com}

\author[SHARIF]{Danial Amini}
\address[SHARIF]{Department of Civil Engineering, Sharif University of Technology, Tehran, Iran}
% \ead{aminidaniel.civil@gmail.com}

\author[MIT]{
\href{https://orcid.org/0000-0002-7370-2332}{\includegraphics[scale=0.06]{orcid.pdf}\hspace{1mm}Ruben Juanes}}
% \ead{juanes@mit.edu}

\cortext[cor1]{Corresponding to: Ehsan Haghighat (ehsanh@mit.edu)}

\hypersetup{
    colorlinks=true,
    linkcolor=blue,
    filecolor=magenta,      
}

\begin{abstract}
Physics-informed neural networks (PINNs) have received significant attention as a unified framework for forward, inverse, and surrogate modeling of problems governed by partial differential equations (PDEs). Training PINNs for forward problems, however, pose significant challenges, mainly because of the complex non-convex and multi-objective loss function. In this work, we present a PINN approach to solving the equations of coupled flow and deformation in porous media for both single-phase and multiphase flow. To this end, we construct the solution space using multi-layer neural networks. Due to the dynamics of the problem, we find that incorporating multiple differential relations into the loss function results in an unstable optimization problem, meaning that sometimes it converges to the trivial null solution, other times it moves very far from the expected solution. We report a dimensionless form of the coupled governing equations that we find most favourable to the optimizer. Additionally, we propose a sequential training approach based on the \emph{stress-split} algorithms of poromechanics. Notably, we find that sequential training based on stress-split performs well for different problems, while the classical \emph{strain-split} algorithm shows an unstable behaviour similar to what is reported in the context of finite element solvers. We use the approach to solve benchmark problems of poroelasticity, including Mandel's consolidation problem, Barry-Mercer's injection-production problem, and a reference two-phase drainage problem. The Python-SciANN codes reproducing the results reported in this manuscript will be made publicly available at \href{https://github.com/sciann/sciann-applications}{https://github.com/sciann/sciann-applications}.
\end{abstract}


\end{frontmatter}

% \newpage
% \tableofcontents
% \newpage

\section{Introduction}\label{sec:intro}

Coupled flow and mechanics is an important physical process in geotechnical, biomechanical, and reservoir engineering \cite{armero1992new, armero1999formulation, park1983stabilization, zienkiewicz1988unconditionally,lewis1998finite, schrefler2004multiphase, white2008stabilized, settari1998coupled, thomas2003coupled, khoei2011numerical}. For example, the pore fluid plays a dominant role in the liquefaction of saturated fine-grain geo-structures such as dams and earthslopes, and has been the cause of instability in infrastructures after earthquakes \cite{zienkiewicz1999computational}. In reservoir engineering, it controls ground subsidence due to water and oil production \cite{fredrich2000geomechanical}. Significant research and development have been done on this topic to arrive at stable and scalable computational models that can solve complex large-scale engineering problems \cite{jha2007locally, white2008stabilized, kim2011stability, jha2014coupled}. Like many other classical computational methods, these models suffer from a major drawback for today's needs: it is difficult to integrate data within these frameworks. To integrate data in the computations, there is often a need for an external optimization loop to match the output to sensor observations. This process constitutes a major challenge in real-world applications.

A recent paradigm in deep learning, known as Physics-Informed Neural Networks \cite{IntroductionPINN}, offers a unified approach to solve governing equations and to integrate data. In this method, the unknown solution variables are approximated by feed-forward neural networks and the governing relations, initial and boundary conditions, and data form different terms of the total loss function. The solution, i.e., optimal parameters of the neural networks, are then found by minimizing the total loss function over random sampling points as well as data points. PINN has resulted in excitement on the use of machine learning algorithms for solving physical systems and optimizing their characteristic parameters given data. PINNs have now been applied to solve a variety of problems including fluid mechanics \cite{IntroductionPINN, jin2021nsfnets, 15-THM-2018, wu2018physics, cai2021physicsfluid, rao2020physics, mao2020physics, reyes2021learning}, solid mechanics \cite{haghighat2021physics, rao2021physics, haghighat2021nonlocal, guo2020energy}, heat transfer \cite{cai2021physics,niaki2021physics,zobeiry2021physics}, electro-chemistry \cite{pilania2018physics, liu2021physics, ji2020stiff}, electro-magnetics \cite{fang2019deep, chen2020physics, noakoasteen2020physics}, geophysics \cite{bin2021pinneik, song2021solving, song2021wavefield, waheed2021holistic}, and flow in porous media \cite{15-PINN-Hamdi-limitations, 43-PINN-similar-Hamdi,41-PINNTerzaghi-forward-inverse,42-PINN-co2-site-response,24-PINN-Biot-forward-inversion,36-PINN-BarryMercer, tartakovsky2020physics} (for a detailed review, see \cite{39-PINN-TotalReview}). A few libraries have also been developed for solving PDEs using PINNs, including SciANN \cite{haghighat2021sciann}, DeepXDE \cite{lu2021deepxde}, SimNet \cite{hennigh2021nvidia}, and NeuralPDE \cite{zubov2021neuralpde}. 

It has been found, however, that training PINNs is slow and challenging for complex problems, particularly those with multiple coupled governing equations, and requires a significant effort on network design and hyper-parameter optimization, often in a trial and error fashion. One major challenge is associated with the multiple-objective optimization problem which is the result of multiple terms in the total loss function. There have been a few works trying to address this challenge. \citet{wang2020understanding} analyzed the distribution of the gradient of each term in the loss function, similar to \citet{chen2018gradnorm}, and found that a gradient scaling approach can be a good choice for specifying the weights of the loss terms. They also later proposed another approach based on the eigenvalues of the neural tangent kernel (NTK) matrix \cite{wang2020ntk} to find optimal weights for each term in the loss function. Another challenge was found to be associated with the partial differential equation itself. For instance, \citet{15-PINN-Hamdi-limitations} reported that solving a nonlinear hyperbolic PDE results in exploding gradients; a regularization was found to provide a solution to the problem. \citet{ji2020stiff} found that solving stiff reaction equations is problematic with PINNs and they addressed that by recasting the fast-reacting relation at equilibrium. Another source of challenges seems to be associated with the simple feed-forward network architecture. \citet{10-PINN-FourierNetwork} proposed a unique network architecture by enriching the input features using Fourier features and the network outputs by a relation motivated by the Fourier series. \citet{haghighat2021nonlocal} reported challenges when dealing with problems developing sharp gradients. They proposed a nonlocal remedy using the Peridynamic Differential Operator \cite{madenci2016peridynamic}. Another remedy has been through space or time discretization of the domain using domain decomposition approaches \cite{5-PINN-cPINN, 33-PINN-XPINN, niaki2021physics}. 

At the time of drafting this work, the studies on the use of PINNs for solving coupled flow-mechanics problems remain uncommon. In a recent work, \citet{41-PINNTerzaghi-forward-inverse} used PINNs to solve Terzaghi's consolidation problem. Terzaghi's problem, however, is effectively decoupled and therefore the authors only need to solve the transient flow problem. \citet{15-PINN-Hamdi-limitations} and \citet{14-PINN-Hamdi-GAN-inversion} studied the Buckley-Leverett problem, where they also only solved the transport problem. \citet{36-PINN-BarryMercer} tried to calibrate the Barry-Mercer problem and reported poor performance of PINNs for the inverse problem. Although it is found that PINNs can work well for low-dimensional inverse problems, using PINNs to solve the forward problems has proven to be very challenging, particularly for coupled flow and mechanics. 

In this work, we study the application of PINNs to coupled flow and mechanics, considering fully saturated (single-phase) and partially saturated (two-phase) laminar flow in porous media. We find that expressing the problem in dimensionless form results in much more stable behavior of the PINN formalism. We study different adaptive weighting strategies and propose one that is most suitable for these problems. We analyze and compare two strategies for training of the feed-forward neural networks: one based on simultaneous training of the coupled equations and one based on sequential training. While being slower, we find that the sequential training using the fixed  stress-split \cite{kim2011stability} strategy results in the most stable and accurate training approach. We then apply the developed framework, in its general form, to solve Mandel's consolidation problem, Barry-Mercer's injection-production problem as well as a two-phase drainage problem. All these problems are developed using the SciANN library \cite{haghighat2021sciann}. This work is the first comprehensive strudy on solving the coupled flow-mechanics equations using PINNs.

\section{Governing Equations}\label{sec:governings}

\subsection{Balance Laws}

The classical formulation of flow in porous media, in which all constituents are represented as a continuum, are expressed as follows. Let us denote the domain of the problem as $\Omega$ with its boundary represented as $\partial\Omega$. The first conservation relation is the mass balance equation, expressed as 
\begin{align}
    \frac{dm_{\alpha}}{dt} + \divg\bs{w}_{\alpha} = \rho_{\alpha}f_{\alpha}, \label{eqs:mass}
\end{align}
where the accumulation term $dm_{\alpha}/dt$ describes the temporal variation of the mass of fluid phase $\alpha$ relative to the motion of the solid skeleton, $\bs{w}_{\alpha}$ is the mass flux of fluid phase $\alpha$ relative to the solid skeleton, and $f_{\alpha}$ is the volumetric source term for phase $\alpha$. 

The second conservation law, under quasi-static assumptions, is the linear momentum balance of solid-fluid mixture expressed as 
\begin{align}
    \divg\bs{\sigma} + \rho_b g \bs{d}=\bs{0}, \label{eqs:linear-momentum}
\end{align}
where $\bs{\sigma}$ is the total Cauchy stress tensor, $\rho_b$ is the bulk density, $g$ is the gravity field directed in $\bs{d}$. 

Conservation of angular momentum results in the symmetry condition on the Cauchy stress tensor as $\bs{\sigma}^T = \bs{\sigma}$. The response of a porous medium is therefore governed by the coupled conservation laws \eqref{eqs:mass} and \eqref{eqs:linear-momentum}. 

\subsection{Small Deformation Kinematics}\label{sec:kinematics}
The total small-strain tensor is defined as 
\begin{align}
    \bs{\varepsilon} = (\grad \bs{u} + \grad\bs{u}^T)/2, \label{eqs:single-phase-kinematics}
\end{align}
with $\bs{u}$ as the displacement vector field of the solid skeleton. The total strain tensor $\bs{\varepsilon}$ is often decomposed into its volumetric and deviatoric parts, i.e., $\varepsilon_v$ and $\bs{e}$, respectively, expressed as 
\begin{align}
    \varepsilon_v &= \tr(\bs{\varepsilon}), \\
    \bs{e} &= \bs{\varepsilon} - \frac{\varepsilon_v}{3} \bs{1}.
\end{align}

\subsection{Single-Phase Poromechanics}\label{sec:single-phase}
For single phase flow in a porous medium, the fluid mass conservation relation reduces to 
\begin{align}
    \frac{dm}{dt} + \divg\bs{w} = \rho_f f, \label{eqs:single-phase-mass-balance}
\end{align}
where $\bs{w}=\rho_f\bs{v}$ is the fluid mass flux, and $\bs{v}$ is the relative seepage velocity, expressed by Darcy's law as 
\begin{align}
    \bs{v} = -\frac{k}{\mu} \left( \grad p - \rho_f g \bs{d} \right), \label{eqs:signle-phase-darcy}
\end{align}
with porous medium's intrinsic permeability as $k$, fluid viscosity as $\mu$, and fluid pressure as $p$. $m$ in \cref{eqs:single-phase-mass-balance} is the mass of the fluid stored in the pores and evaluated as 
\begin{align}
    m = \rho_f \phi(1+\varepsilon_v)
\end{align}
with $\phi$ as the porosity of the medium, expressed as the ratio of void volume to the bulk volume as $\phi=V_p/V_b$. The change in the fluid content is governed in the theory of poroelasticity by  
\begin{align}
    \frac{\delta m}{\rho_{f}} = b~\varepsilon_v + \frac{1}{M} \delta p, \label{eqs:single-phase-fluid-content}
\end{align}
where $M$ and $b$ are the Biot bulk modulus and the Biot coefficient, and are related to fluid and rock properties as 
\begin{align}
    \frac{1}{M} = \phi_0 c_f + \frac{b-\phi_0}{K_s}, \quad b=1-\frac{K_{dr}}{K_s}. \label{eqs:single-phase-biot-params}
\end{align}
Here, $c_f$ is the fluid compressibility defined as $c_f = 1/K_f$, $K_f$ and $K_s$ are the bulk moduli of fluid and solid phases respectively, and $K_{dr}$ is the drained bulk modulus. 

As expressed by Biot \cite{biot1941general}, the poroelastic response of the bulk due to changes in the pore pressure is expressed as 
\begin{align}
    \delta \bs{\sigma} = \bmc{C}_{dr} : \bs{\varepsilon} - b~\delta p \bs{1}, \label{eqs:single-phase-biot-elasticity} 
\end{align}
where, $\bmc{C}_{dr}$ is the 4th order drained elastic stiffness tensor and $\bs{1}$ is the 2nd order identity tensor. This implies that the effective stress, responsible for the deformation in the soil, can be defined as  $\delta\bs{\sigma}' = \delta\bs{\sigma} + b~\delta p \bs{1}$. The isotropic elasticity stress-strain relations implies that $\bmc{C}_{dr} = K_{dr}\bs{1}\dyad\bs{1} + 2G(\mathbb{I} - \frac{1}{3}\bs{1}\dyad\bs{1})$, with $\mathbb{I}$ as the fourth-order identity tensor and $G$ as the shear modulus. For convenience, we can express $G$ as a function of Poisson ratio $\nu$ and the drained bulk modulus as 
\begin{align*}
    G = \frac{3}{2} \nu^* K_{dr}, \quad \nu^* = \frac{1-2\nu}{1+\nu}. 
\end{align*}
We can therefore write, 
\begin{align}
    \delta\bs{\sigma} &= K_{dr}\varepsilon_v\bs{1} + 3\nu^*K_{dr} \bs{e} - b\delta p \bs{1}. \label{eqs:single-phase-stress-inc}
\end{align}
The volumetric part of the stress tensor, also known as the hydrostatic invariant, is defined as $\sigma_v = \tr(\bs{\sigma})/3$, which can be expressed as 
\begin{align}
    \delta \sigma_v = K_{dr} \varepsilon_v - b~\delta p. \label{eqs:single-phase-volumetric-stress}
\end{align}
The pressure field is coupled with the stress field through the volumetric strain. Since, neural networks have multi-layer compositional mathematical forms, with their derivative taking very complex forms, instead of evaluating the volumetric strain from displacements, we keep it as a separate solution variable subject to the following kinematic law 
\begin{align}
    \varepsilon_v - \divg{\bs{u}} = 0.
\end{align}

Therefore, the final form of the relations describing the problem of single-phase flow in a porous medium are derived by expressing the poroelasticity constitutive relations as 
\begin{align}
    & \bs{\sigma} - \bs{\sigma}_0 = K_{dr}\varepsilon_v\bs{1} + 3\nu^*K_{dr}\bs{e} - b(p-p_0) \bs{1}, \label{eqs:single-phase-total-stress}\\
    & \frac{1}{\rho_{f,0}}(m-m_0) = b \varepsilon_v + \frac{1}{M} (p - p_0). \label{eqs:single-phase-total-fluid-content}
\end{align}
We can then write the linear momentum relation \eqref{eqs:linear-momentum} in terms of displacements as 
\begin{align}
    K_{dr} \grad \varepsilon_v + \frac{1}{2} \nu^* K_{dr} \grad(\divg \bs{u}) + \frac{3}{2}\nu^*K_{dr} \divg (\grad\bs{u}) - b~\grad p + \rho_b g \bs{d} = \bs{0}. \label{eqs:navier}
\end{align}
Note that for reasons we will discuss later, we do not merge the $\varepsilon_v$ and $\divg{\bs{u}}$ terms in this relation. Combining \cref{eqs:single-phase-mass-balance,eqs:signle-phase-darcy,eqs:single-phase-fluid-content} and after some algebra, we can express the mass balance relation in terms of the volumetric strain as 
\begin{align}
    \frac{1}{M} \partt{p} + b \partt{\varepsilon_v} + \divg \bs{v} = f, \label{eqs:single-phase-mass-conservation-volstrain}
\end{align}
or, equivalently, in terms of the volumetric stress as 
\begin{align}
    \left(\frac{b^2}{K_{dr}} + \frac{1}{M}\right) \partt{p} + \frac{b}{K_{dr}} \partt{\sigma_v} + \divg \bs{v} = f. \label{eqs:single-phase-mass-conservation-volstress}
\end{align}

\paragraph{\textbf{Dimensionless relations}}
Since the deep learning optimization methods work most effectively on dimensionless data, we derive the dimensionless relations here. Let us consider the dimensionless variables, denoted by $\bar{\circ}$, as 
\begin{align}
    \bar{t} = \frac{t}{t^*}, ~
    \bar{\bs{x}} = \frac{\bs{x}}{x^*}, ~
    \bar{\bs{u}} = \frac{\bs{u}}{u^*}, ~
    \bar{\bs{\varepsilon}} = \frac{\bs{\varepsilon}}{\varepsilon^*}, ~
    \bar{p} = \frac{p}{p^*}, ~
    \bar{\bs{\sigma}} = \frac{\bs{\sigma}}{p^*}, ~
    \label{eqs:single-phase-dimensionless-coords}
\end{align}
where $t^*, x^*, u^*, \varepsilon^*, p^*$ are dimensionless factors to be defined later. This implies that the partial derivatives and the divergence operator are also expressed as 
\begin{align}
    \partt{\circ} = \frac{1}{{t^*}} \parttd{\circ}, \quad 
    \divg{\circ} = \frac{1}{x^*} \divgd{\circ}, \quad
    \label{eqs:single-phase-dimensionless-derivatives}
\end{align}
and therefore, we can express the seepage velocity as 
\begin{align}
    \bs{v} = -\frac{k}{\mu}\frac{p^*}{x^*} (\gradd \bar{p} - N_d \bs{d}). \label{eqs:single-phase-dimensionless-seepage}
\end{align}
The Navier displacement and stress-strain relations are expressed in the dimensionless form as
\begin{align}
    & \gradd \bar{\varepsilon_v} + \frac{1}{2}\nu^* \gradd (\divgd \bar{\bs{u}}) + \frac{3}{2}\nu^* \divgd (\gradd  \bar{\bs{u}}) - b \gradd \bar{p} + N_d \bs{d} = \bs{0}, \label{eqs:nondim1}\\
    & \bar{\bs{\sigma}} - \bar{\bs{\sigma}}_0 = \bar{\varepsilon}_v \bs{1} + 3\nu^* \bar{\bs{e}} - b(\bar{p} - \bar{p}_0) \bs{1}, \label{eqs:nondim2}\\
    & \delta\bar{\sigma}_v = \bar{\varepsilon}_v - b~\delta\bar{p}.\label{eqs:nondim3}
\end{align}
The dimensionless form of \cref{eqs:single-phase-mass-conservation-volstrain,eqs:single-phase-mass-conservation-volstress} is expressed, in terms of stress, as 
\begin{align}
    & \parttd{\bar{p}} - \gradd^2 \bar{p} + D^* \parttd{\bar{\sigma}_{v}}  = f^*, \label{eqs:nondim4}
\end{align}
or, in terms of strain, as
\begin{align}
    & (1-bD^*)\parttd{\bar{p}} - \gradd^2 \bar{p} + D^* \parttd{\bar{\varepsilon}_{v}}  = f^*, \label{eqs:nondim5}
\end{align}
where the dimensionless parameters are
\begin{equation}\label{eqs:nondim6}
\begin{split}
   &u^* = \frac{p^*}{K_{dr}} x^*,~
    \varepsilon^* = \frac{u^*}{x^*},~
    N_d = \frac{\rho_b g x^*}{p^*},~\\
   &t^* = \left( \frac{b^2}{K_{dr}} + \frac{1}{M} \right) \frac{\mu}{k} x^{*2},~
    D^* = \frac{b M}{b^2 M + K_{dr}},~
    f^* = \frac{\mu}{k}\frac{x^{*2}}{p^*} f.
\end{split}
\end{equation}
In sequential iterations, \cref{eqs:nondim4} is known as the \emph{fixed-stress-split} approach, while the use of \cref{eqs:nondim5} is referred to the \emph{fixed-strain-split} method \cite{kim2011stability}.

%\newpage
\subsection{Two-Phase Flow Poromechanics}
In the case of multiphase flow poromechanics, let us first define the `equivalent pressure', expressed as 
\begin{align}
  {p_E} = \sum_{\alpha}{S_\alpha}{p_\alpha} - {U},
\end{align}
with $S_{\alpha}$ and $p_{\alpha}$ as the degree of saturation and pressure of the fluid phase $\alpha$, respectively, and ${U}$ as the interfacial energy between phases, defined incrementally as ${\delta{U} = \sum_{\alpha}{p_\alpha}{\delta S_\alpha}}$ \cite{kim2013rigorous, coussy2004poromechanics}. Let us also define here the capillary pressure $p_c$, denoting the pressure difference between the nonwetting fluid phase $n$ and the wetting fluid phase $w$ as 
\begin{align}
{p_{c}} = {p_{n}} - {p_w}.
\end{align}

The poroelastic stress-strain relation of a multiphase system is expressed as
\begin{align}
  {\delta {\bs{\sigma}}} &= K_{dr}\varepsilon_v\bs{1} + 3\nu^*K_{dr}\bs{e} - {b}~{\delta p_E}{\bs{1}}, \\
  \delta \sigma_v &= K_{dr}\varepsilon_v - b\delta p_E.
\end{align}
where $b\delta p_E = \sum_\alpha b_\alpha \delta p_\alpha$. The mass of the fluid phase $\alpha$ is expressed as 
\begin{align}
  {m_\alpha} = {{\rho_\alpha}{S_\alpha}{\phi}({1 + \varepsilon_v})}. \label{eqs:fluid_mass_multi}
\end{align}
We can now express the fluid content for phase $\alpha$ as 
\begin{align}
  \frac{dm_\alpha}{\rho_\alpha} = b_\alpha d\varepsilon_v + \sum_{\beta} N_{\alpha \beta} dp_\beta. \label{eqs:fluid_mass_inc_multi}
\end{align}
Substituting \eqref{eqs:fluid_mass_inc_multi} into \eqref{eqs:mass}, we can write the fluid mass conservation equation in terms of volumetric strain as:
\begin{align}
\sum_{\beta}{{N_{\alpha\beta}}}\frac{\partial p_\beta}{\partial t} + {b_\alpha}\frac{\partial \varepsilon_v}{\partial t} + {\divg {\bs{v_\alpha}}} = {f_\alpha},
\end{align}
where ${b_\alpha} = {S_\alpha}{b}$ and ${N_{\alpha\beta}}$ expresses components of the inverse Biot modulus matrix for a multiphase system which can be obtained from fluid mass equation for two-phase flow system as:
\begin{align}
  & {N_{nn}} = {-\phi}\frac{\partial {S_w}}{\partial p_{c}} + {\phi}{S_{n}}{c_{n}} + {S_{n}^2}{N}, \\
  & {N_{nw}} ={N_{wn}} = {\phi}\frac{\partial {S_w}}{\partial p_{c}} + {S_{n}}{S_w}{N}, \\
  & {N_{ww}} = {-\phi}\frac{\partial {S_w}}{\partial p_{c}} + {\phi}{S_w}{c_w} + {S_w^2}{N},
\end{align}
with ${{N} = ({b - \phi}){(1 - b)}/{K_{dr}}}$. The seepage velocity of the fluid phase $\alpha$, i.e., ${\bs{v_\alpha}}$, is obtained by extending the Darcy's law as 
\begin{align}
 {\bs{v_\alpha}} = -\frac{k}{\mu}{k_r^\alpha}({\grad p_\alpha}-{\rho_\alpha}{g}\mc{\bs{d}}),
\end{align}
where ${{\rho_\alpha}{g}\mc{\bs{d}}}$ denotes the gravity force vector for the phase ${\alpha}$, and ${k_r^\alpha}$ is the relative permeability of phase ${\alpha}$ . Considering the relation between the volumetric strain and equivalent volumetric stress,
\begin{align}
\frac{b_\alpha}{K_{dr}}\frac{\partial \sigma_v}{\partial t} = {b_\alpha}\frac{\partial \varepsilon_v}{\partial t} - \sum_{\beta}\frac{b_\alpha b_\beta}{K_{dr}}\frac{\partial p_\beta}{\partial t},
\end{align}
Consequently, the fluid mass conservation equation can be rewritten in terms of equivalent volumetric stress for phase ${\alpha}$ \cite{kim2013rigorous, jha2014coupled},
\begin{align}
\sum_{\beta}({{N_{\alpha\beta}} + \frac{b_\alpha b_\beta}{K_{dr}}})\frac{\partial p_\beta}{\partial t} + \frac{b_\alpha}{K_{dr}}\frac{\partial \sigma_v}{\partial t} + \divg {\bs{v_\alpha}} = {f_\alpha},
\end{align}

\paragraph{\textbf{Dimensionless relations}}
Here, we define dimensionless variables for two-phase poromechanics as:
\begin{align}
\bar{t} =\frac{t}{t^*}\space , \space \bar{x} = \frac{x}{x^*}\space , \space \bar{p}_w =\frac{p_w}{p^*}\space , \space \bar{p}_{n} = \frac{p_{n}}{p^*}\space , \space \bar{\bs{\sigma}} = \frac{\bs{\sigma}}{p^*}\space , \space \bs{\bar{u}} = \frac{\bs{u}}{u^*}\space , \space \bar{\bs{\varepsilon}} = \frac{\bs{\varepsilon}}{\varepsilon^*}.
\end{align}
Noting that ${\mu = \mu_{n} + \mu_w}$, and considering the dimensionless parameters 
\begin{equation}
\begin{split}
    & \frac{1}{\bar{M}} = {\phi}{S_w}{c_w} + {\phi}{S_{n}}{c_{n}} + \frac{b - \phi}{K_s}, ~~
     \frac{1}{\bar{M}^*} = \frac{1}{\bar{M}} + \frac{b^2}{K_{dr}}, \\
    & {t^*} = \frac{\mu {x^*}^2}{k \bar{M}^*}, ~~
      {D_\alpha^*} = {S_\alpha}\frac{b \bar{M}^*}{K_{dr}}, ~~
      {N_d} = \frac{{x^*} {\rho}}{p^*}{g}, ~~
      {f_\alpha^*} = {f_\alpha}\frac{\mu {{x^*}^2}}{k {p^*}}, \\
    & {u^*} = \frac{p^*}{K_{dr}}{x^*}, ~~
      {\bar{\bs{\varepsilon}}} = \frac{\bs{\varepsilon}}{\varepsilon^*}, ~~
      \varepsilon^* = \frac{u^*}{x^*},
\end{split}
\end{equation}   
the dimensionless \emph{strain-split} fluid mass conservation equation takes the form
\begin{align}
 & \sum_{\beta}{N_{\alpha\beta}}{\bar{M}^*}\frac{\partial \bar{p}_{\beta}}{\partial \bar{t}} + {D_\alpha^*}\frac{\partial \bar{\varepsilon}_v}{\partial \bar{t}} - \frac{\mu}{\mu_\alpha} {\divgd[{k_r^\alpha}({\gradd{\bar{p}_\alpha}}- \frac{\rho_\alpha}{\rho}{N_d}\mc{\bs{d}})]} = {f_\alpha^*}, \label{eqs:multi phase fluid mass epsilon}
\end{align}
and the dimensionless \emph{stress-split} formulation takes the form 
\begin{align}
    & \sum_{\beta}({N_{\alpha\beta}} + \frac{b_\alpha b_\beta}{K_{dr}}){\bar{M}^*}\frac{\partial \bar{p}_{\beta}}{\partial \bar{t}} + {D_\alpha^*}\frac{\partial \bar{\sigma}_v}{\partial \bar{t}} - \frac{\mu}{\mu_\alpha} {\divgd[{k_r^\alpha}({\gradd{\bar{p}_\alpha}}- \frac{\rho_\alpha}{\rho}{N_d}\mc{\bs{d}})]} = {f_\alpha^*},
\end{align}
in which ${\rho = (1 - \phi)\rho_s + \frac{1}{2}\phi({\rho_w + \rho_n})}$. The linear momentum equation for the whole medium and the stress-strain relation in terms of dimensionless variables are expressed as 
\begin{align}
    & \gradd \bar{\varepsilon}_v +  \frac{1}{2}\nu^* \gradd (\divgd \bar{\bs{u}}) + \frac{3}{2}\nu^* \divgd (\gradd  \bar{\bs{u}}) - {b} \sum_{\beta}{\gradd{S_\beta\bar{p}_\beta}} + \frac{\rho_b}{\rho}{N_d}\mc{\bs{d}} = \bs{0}, \\
    & \frac{\partial \bar{\bs{\sigma}}}{\partial \bar{t}} = \frac{\partial \bar{\varepsilon}_v}{\partial \bar{t}} \mathbf{1} + 3{\nu^*}\frac{\partial \bar{\bs{e}}}{\partial \bar{t}} - {b}{\sum_{\beta}{S_\beta}\frac{\partial {\bar{p}}_\beta}{\partial \bar{t}}\mathbf{1}},
   \end{align}
where ${\rho_b=(1-\phi){\rho_s}+\phi{S_w}{\rho_w}+\phi{S_n}{\rho_n}}$ is the bulk density.

\section{Physics-Informed Neural Networks}

We approximate the solution variables of the poromechanics problem using deep neural networks. The governing equations and initial/boundary conditions form different terms of the loss function. The total loss (error) function is then defined as a linear composition of these terms \cite{IntroductionPINN}. The optimal weights of this composition are either found by trial and error or through more fundamental approaches such as gradient normalization that will be discussed later \cite{IntroductionPINN,wang2020understanding}. Solving the problem, i.e., identifying the parameters of the neural network, is then done by minimizing the total loss function on random collocation points inside the domain and on the boundary. It has been shown that minimizing such a loss function is a non-trivial optimization task, particularly for the case of coupled differential equations (see \cite{39-PINN-TotalReview}), and special treatment is needed, which we will discuss later in this text. 

Let us define the nonlinear mapping function $\Sigma$ as 
\begin{align}
    \hat{\bs{y}}^i = \Sigma^i(\hat{\bs{x}}^{i-1}) := \sigma^i(\mathbf{W}^i \hat{\bs{x}}^{i-1} + \mathbf{b}^i), \quad i=1,2,\dots,L, \label{eqs:nn_layer}
\end{align}
where, $\hat{\bs{x}}^{i-1} \in \mathbb{R}^{I_i}$ and $\hat{\bs{y}}^i \in \mathbb{R}^{O_i}$ are considered the input and output of hidden layer $i$, with dimensions $I_i$ and $O_i$, respectively. {For the first layer, i.e. , $i=1$, $\hat{\bs{x}}^{i-1} = \bs{x}$, and for other layers, $\hat{\bs{x}}^{i-1}$ is the output of the previous layer. $L$ represents total number of transformations {\cref{eqs:nn_layer}}, a.k.a. layers, that should be applied to the inputs to reach the final output in a \emph{forward pass}.} The parameters of this transformation $\mathbf{W}^i \in \mathbb{R}^{O_i\times I_i}$ and $ \mathbf{b}^i \in \mathbb{R}^{O_i}$ are known as weights and biases, respectively. The activation function $\sigma^i: \mathbb{R} \rightarrow \mathbb{R}$ introduces nonlinearity to the transformation. For PINNs, $\sigma^i$ is often taken as the \emph{hyperbolic-tangent} function and kept the same for all hidden layers. With this definition as the main building block of a neural network, we can now construct an $L$-layer feed-forward neural network as
\begin{align}
    y = \mathcal{N}(\bs{x}, t; \bs{\theta}) := \Sigma^L\circ\Sigma^{L-1}\circ \dots \circ \Sigma^{1}(\bs{x}, t),
\end{align}
where $\circ$ represent the composition operator, and $\bs{\theta}\in\mathbb{R}^M$ as the collection of all $M$ parameters of the network to be identified through the optimization, with $M=\sum_{i=1}^{L} (O_i\times I_i + O_i)$. The main input features of this transformation are spatio-temporal coordinates $\bs{x}\in\mathbb{R}^{D}$ and $t\in\mathbb{R}$, with $D$ as the spatial dimension, and the ultimate output is the network output $y\in\mathbb{R}$. Note that the last layer (output layer) is often kept linear for regression-type problems, i.e., $\sigma^L(\bs{z}) = \bs{z}$. In the context of PINNs, the output can be the desired solution variables such as displacement or pressure fields. 

Finally, let us consider a time-varying partial differential equation $\mc{P} u(\bs{x}, t) = f(\bs{x},t)$ with $\mc{P}$ as the differential operator and $f(\bs{x},t)$ as the source term, defined in the domain $\Omega\times T$, with $\Omega \in \mathbb{R}^D$ and $T\in\mathbb{R}$ representing the spatial and temporal domains, respectively. This PDE is subjected to Dirichlet boundary condition $g_D(\bs{x}, t)$ on $\Gamma_D$, Neumann boundary condition $g_N(\bs{x}, t)$ on $\Gamma_N$, where $\Gamma_D \cup \Gamma_N = \partial\Omega$ and $\Gamma_D \cap \Gamma_N = \varnothing$. It is also subject to the initial condition $h(\bs{x})$ at $T_0$. Based on the PINNs, the unknown variable $u(\bs{x},t)$ is approximated using a multi-layer neural network, i.e., $u(\bs{x},t) \approx \tilde{u}(\bs{x},t) = \mc{N}_u(\bs{x}, t; \bs{\theta})$. The PDE residual $\mc{P} \tilde{u} - f$ is evaluated using automatic differentiation. With $\mathbf{n}$ as the outward normal to the boundary, the boundary flux term is evaluated as $\tilde{q} = \mathbf{n}\cdot \grad{\tilde{u}}$. Therefore, the residual terms for the initial and boundary conditions are constructed accordingly. The total loss function is then expressed as 
\begin{equation}
\begin{split}
    \mc{L}(\bs{x}, t; \bs{\theta}) &= \lambda_1 \left\| \mc{P}\tilde{u}(\bs{x},t) - f(\bs{x},t) \right\|_{\Omega\times T} \\
    &+ \lambda_2 \left\|\tilde{u}(\bs{x},t) - g_D(\bs{x},t) \right\|_{\Gamma_D\times T} \\
    &+ \lambda_3 \left\|\tilde{q}(\bs{x},t) - g_N(\bs{x},t) \right\|_{\Gamma_N\times T} \\
    &+ \lambda_4 \left\|\tilde{u}(\bs{x},t_0) - h(\bs{x}) \right\|_{\Omega\times T_0}
\end{split}
\end{equation}
where $\lambda_i$ represent the weight (penalty) of each term constructing the total loss function. The solution to this initial and boundary value problem is found by minimizing the total loss function on a set of collocation points $\mathbf{X}\in\mathbb{R}^{N\times D}, \mathbf{T}\in\mathbb{R}^{N\times 1}$ with $N$ as the total number of collocation points. This optimization problem is expressed mathematically as
\begin{align}
\bs{\theta}^* = \argmin_{\bs{\theta}\in\mathbb{R}^D} \mc{L}(\mathbf{X}, \mathbf{T}; \bs{\theta}),
\end{align}
with $\bs{\theta}^*$, as the \emph{optimal} values for the network parameters, minimizing the total loss function above \cite{IntroductionPINN}. 

\subsection{Optimization method}
The most common approach for training neural networks in general and PINNs in particular is using the Adam \cite{kingma2014adam} optimizer from the first-order stochastic gradient descent family. This method is highly scalable and therefore has been successfully applied to many supervised-learning tasks. For the case of multi-objective optimization, which arise with PINNs, fine-tuning its learning rate and weights associated with the individual loss terms is challenging. 

Another relatively common approach for training PINNs is the second-order BFGS method and its better memory-efficient variant, L-BFGS\cite{liu1989limited}; this method, however, lacks scalability and becomes very slow for deep neural networks with large number of parameters. It does not have some drawbacks of the first order methods particularly the learning rate is automatically identified through the Hessian matrix \cite{nocedal2006numerical}. In this study, we use the Adam optimizer with an initial learning rate of $10^{-3}$ to train our networks. 

\subsection{Adaptive weights}
Another challenge in training PINNs is associated with the gradient pathology of each loss term in the total loss function \cite{chen2018gradnorm}. If we look at the gradient vector of each loss term with respect to network parameters, we can find that the terms with higher derivatives tend to have larger gradients, and therefore dominate more the total gradient vector used to train the network \cite{wang2020understanding}. This is however not desirable in PINNs since the solution to the PDEs depends heavily on the accurate imposition of the boundary conditions. 

To address this, \citet{chen2018gradnorm} proposed a gradient scaling algorithm, known as \emph{GradNorm}, as described below. Let us consider the total loss function is expressed as $\mc{L} = \sum_i \lambda_i \mc{L}_i$ with $\mc{L}_i$ as different loss terms such as the PDE residual or boundary conditions and $\lambda_i$ as their weights. Let us also define the total gradient vector $\bs{\mc{G}}$ as the gradient of total loss with respect to network parameters $\bs{\theta}$, which can be expressed as $\bs{\mc{G}} = \sum_i \bs{\mc{G}}_i$, with $\bs{\mc{G}}_i$ as the gradient vector of the weighted loss term $i$ with respect to network parameters $\bs{\theta}$, i.e., $\bs{\mc{G}}_i= \nabla_{\bs{\theta}} (\lambda_i\mc{L}_i)$, and with $\mc{G}_i = \left\|\bs{\mc{G}}_i\right\|_2$ as its $L^2$-norm. According to this approach, the objective is to find $\lambda_i$s such that the $L^2$ of the gradient vector of the weighted loss term, i.e., $\mc{G}_i^2$, approaches the average $L^2$ norm of all loss terms, factored by a score value, i.e., 
\begin{align}
    \mc{G}_i \rightarrow E_{\forall j} \left[ \mc{G}_j\right]~s_i^\alpha, \quad \text{where} \quad s_i = \frac{\tilde{\mc{L}}_i}{E_{\forall j} \left[ \mc{L}_j \right]}, \quad \tilde{\mc{L}}_i = \frac{\mc{L}_i}{\mc{L}_i(0)},
\end{align}
with, $\mc{L}_i(0)$ as the value of loss term $\mc{L}_i$ at the beginning of training, $\tilde{\mc{L}}_i$ as the relative value for the loss term, $s_i$ as a measure of how much a term is trained compared to all other terms, referred here as the \emph{score} of loss term $\mc{L}_i$. To find the weights, the authors propose a gradient descent update to minimize the  \emph{GradNorm} loss function, as 
\begin{align}
\lambda_i^{\tau} = \lambda_i^{\tau-1} - \beta \frac{\partial \mc{L}_{GN}}{\partial \lambda_i}, \quad \text{where} \quad
    \mc{L}_{GN} = \sum_i \left|\mc{G}_i - E_{\forall j} \left[ \mc{G}_j\right]~s_i^\alpha\right|.
\end{align}
Knowing that the minimum of summation of positive functions is zero and occurs when all functions are zero, we can algebraically recast the weights $\hat{\lambda}_i$ such that it zeros the $\mc{L}_{GN}$. Then using an Euler update, as proposed by \citet{wang2020understanding}, the loss weight $\lambda_i$ at training epoch $\tau$ can be expressed as 
\begin{align}
\lambda_i^{\tau} = (1-\beta) \lambda_i^{\tau-1} + \beta \hat{\lambda}_i^{\tau}, \quad \text{where} \quad
\hat{\lambda}_i^\tau = \lambda^{\tau-1}_i \frac{E_{\forall j} \left[ \mc{G}_j\right]}{\mc{G}_j}~s_i^\alpha.
\end{align}
We therefore use loss-term weights evaluated following this strategy. There are also adaptive spatial and temporal sampling strategies offered in the literature \cite{nabian2021efficient, wight2020solving}; however, we do not employ them in this work.

\section{PINN-PoroMechanics}\label{sec:pinn-poromechanics}

Now that we discussed the method of Physics-Informed Neural Networks (PINNs), we can express the PINN solution strategy for the single-phase and multiphase poroelasticity problem discussed in \cref{sec:governings}. For definiteness, we consider two-dimensional problems. 

\subsection{PINN solution of single-phase poroelasticity}
For the case of single-phase poroelasticity, the unknown solution variables are $u_x, u_y, \text{and}~ p$. Considering that $p$ is correlated with the volumetric strain $\varepsilon_v$, and also considering that the derivatives of multi-layer neural network takes very complicated forms, we also take the volumetric strain as an unknown so that the we can better enforce the coupling between fluid pressure and displacements. This results in an additional conservation PDE on the volumetric strain, expressed as
\begin{align}
    \bar{\varepsilon}_v - \divgd \bar{\bs{u}} = 0.
\end{align}
By trial and error, we find this strategy to improve the training; 
% there is more theoretical foundations for it within the FEM community. 
This strategy of explicitly imposing volumetric constrains has a long history in FEM modeling of quasi-incompressible materials \cite{zienkiewicz1977finite,hughes2012finite}.
Therefore, the neural networks for the dimensionless form of these variables are,
\begin{align}
    \bar{u}_{\bar{x}} &: (\bar{x}, \bar{y}, \bar{t}) \mapsto \mathcal{N}_{\bar{u}_{\bar{x}}}(\bar{x}, \bar{y}, \bar{t}; \bs{\theta}_{\bar{u}_{\bar{x}}}), \\
    \bar{u}_{\bar{y}} &: (\bar{x}, \bar{y}, \bar{t}) \mapsto \mathcal{N}_{\bar{u}_{\bar{y}}}(\bar{x}, \bar{y}, \bar{t}; \bs{\theta}_{\bar{u}_{\bar{y}}}), \\
    \bar{p} &: (\bar{x}, \bar{y}, \bar{t}) \mapsto \mathcal{N}_{\bar{p}}(\bar{x}, \bar{y}, \bar{t}; \bs{\theta}_{\bar{p}}), \\
    \bar{\varepsilon}_v &: (\bar{x}, \bar{y}, \bar{t}) \mapsto \mathcal{N}_{\bar{\varepsilon}_v}(\bar{x}, \bar{y}, \bar{t}; \bs{\theta}_{\bar{\varepsilon}_v}), 
\end{align}
where $\bs{\theta}_\alpha$ highlights that these networks have independent parameters, as proposed in our earlier work \cite{haghighat2021physics}. The total coupled loss function is then expressed as 
\begin{equation}
\begin{split}
\mc{L} &= \lambda_1 \left\| \frac{\partial \bar{\varepsilon}_{v}}{\partial \bar{x}} + \frac{\nu^*}{2}\left(\frac{\partial^2 \bar{u}_{\bar{x}}}{\partial\bar{x}^2} + \frac{\partial^2 \bar{u}_{\bar{y}}}{\partial\bar{x}\partial\bar{y}} \right) + \frac{3\nu^*}{2}\left(\frac{\partial^2 \bar{u}_{\bar{x}}}{\partial\bar{x}^2} + \frac{\partial^2 \bar{u}_{\bar{x}}}{\partial\bar{y}^2} \right) - b\frac{\partial \bar{p}}{\partial \bar{x}} - N_d~d_{\bar{x}} \right\| \\
&+ \lambda_2 \left\| \frac{\partial \bar{\varepsilon}_{v}}{\partial \bar{y}} + \frac{\nu^*}{2}\left(\frac{\partial^2 \bar{u}_{\bar{x}}}{\partial\bar{x}\partial\bar{y}} + \frac{\partial^2 \bar{u}_{\bar{y}}}{\partial\bar{y}^2} \right) + \frac{3\nu^*}{2}\left(\frac{\partial^2 \bar{u}_{\bar{y}}}{\partial\bar{x}^2} + \frac{\partial^2 \bar{u}_{\bar{y}}}{\partial\bar{y}^2} \right) - b\frac{\partial \bar{p}}{\partial \bar{y}} - N_d~d_{\bar{y}} \right\| \\
&+ \lambda_3 \left\|\frac{\partial\bar{p}}{\partial\bar{t}} - \left( \frac{\partial^2 \bar{p}}{\partial \bar{x}^2} + \frac{\partial^2 \bar{p}}{\partial \bar{y}^2} \right) + D^* \frac{\partial\bar{\sigma}_v}{\partial \bar{t}} - f^*\right\| \\
&+ \lambda_4 \left\| \bar{u}_{\bar{x}} - \tilde{u}_{\bar{x}} \right\| + \lambda_5  \left\| \bar{u}_{\bar{y}} - \tilde{u}_{\bar{y}} \right\| + \lambda_6 \left\| \bar{\sigma}_{\bar{xx}} - \tilde{\sigma}_{\bar{xx}} \right\| + \lambda_7 \left\| \bar{\sigma}_{\bar{yy}} - \tilde{\sigma}_{\bar{yy}} \right\| +\lambda_8 \left\| \bar{\sigma}_{\bar{xy}} - \tilde{\sigma}_{\bar{xy}} \right\| \\
&+ \lambda_9 \left\| \bar{p} - \tilde{p} \right\| + \lambda_{10} \left\| \bar{q}_{\bar{x}} - \tilde{q}_{\bar{x}} \right\| + \lambda_{11}  \left\| \bar{q}_{\bar{y}} - \tilde{q}_{\bar{y}} \right\| \\
&+ \lambda_{12} \left\| \bar{\varepsilon}_v - \left(\frac{\partial \bar{u}_{\bar{x}}}{\partial \bar{x}} + \frac{\partial\bar{u}_{\bar{y}}}{\partial\bar{y}} \right)\right\|, \label{eqs:SimultaneousTraining}
\end{split}
\end{equation}
where $\tilde{\circ}$ implies the boundary or initial values for each term. 
% As you find, there are many terms in the total loss function, which causes immense challenges for the optimizer to train the networks. We experience that without the use of scored-adaptive-weight, we were not able to get any meaningful result by optimizing this loss function using Adam optimizer. 
The many terms in the total loss function, and their complexity, pose enormous challenges in the training of the network. Meaningful results of the Adam optimizer were obtained only with the use of scored-adaptive weights, but our experience has been that the procedure lacks robustness. 
% However, another strategy that worked successfully was based the stress-split sequential training of two separate loss functions for the solid and fluid phases as 
To address the inadequacy of traditional methods of network training to multiphysics problems, we propose a novel strategy, inspired by the success of the fixed-stress operator split for time integration of the forward problem in coupled poromechanics \cite{kim2011stability}. 

We adopt a sequential training of the flow and mechanics sub-problems, expressed in fixed-stress split form:
\begin{equation}
\begin{split}
\mc{L}_f &= \lambda_3 \left\|\frac{\partial\bar{p}}{\partial\bar{t}} - \left( \frac{\partial^2 \bar{p}}{\partial \bar{x}^2} + \frac{\partial^2 \bar{p}}{\partial \bar{y}^2} \right) + D^* \frac{\partial\bar{\sigma}_v}{\partial \bar{t}} - f^*\right\| \\
&+ \lambda_9 \left\| \bar{p} - \tilde{p} \right\| + \lambda_{10} \left\| \bar{q}_{\bar{x}} - \tilde{q}_{\bar{x}} \right\| + \lambda_{11}  \left\| \bar{q}_{\bar{y}} - \tilde{q}_{\bar{y}} \right\|, \\
\mc{L}_s &= \lambda_1 \left\| \frac{\partial \bar{\varepsilon}_{v}}{\partial \bar{x}} + \frac{\nu^*}{2}\left(\frac{\partial^2 \bar{u}_{\bar{x}}}{\partial\bar{x}^2} + \frac{\partial^2 \bar{u}_{\bar{y}}}{\partial\bar{x}\partial\bar{y}} \right) + \frac{3\nu^*}{2}\left(\frac{\partial^2 \bar{u}_{\bar{x}}}{\partial\bar{x}^2} + \frac{\partial^2 \bar{u}_{\bar{x}}}{\partial\bar{y}^2} \right) - b\frac{\partial \bar{p}}{\partial \bar{x}} - N_d~d_{\bar{x}} \right\| \\
&+ \lambda_2 \left\| \frac{\partial \bar{\varepsilon}_{v}}{\partial \bar{y}} + \frac{\nu^*}{2}\left(\frac{\partial^2 \bar{u}_{\bar{x}}}{\partial\bar{x}\partial\bar{y}} + \frac{\partial^2 \bar{u}_{\bar{y}}}{\partial\bar{y}^2} \right) + \frac{3\nu^*}{2}\left(\frac{\partial^2 \bar{u}_{\bar{y}}}{\partial\bar{x}^2} + \frac{\partial^2 \bar{u}_{\bar{y}}}{\partial\bar{y}^2} \right) - b\frac{\partial \bar{p}}{\partial \bar{y}} - N_d~d_{\bar{y}} \right\| \\
&+ \lambda_4 \left\| \bar{u}_{\bar{x}} - \tilde{u}_{\bar{x}} \right\| + \lambda_5  \left\| \bar{u}_{\bar{y}} - \tilde{u}_{\bar{y}} \right\| + \lambda_6 \left\| \bar{\sigma}_{\bar{xx}} - \tilde{\sigma}_{\bar{xx}} \right\| + \lambda_7 \left\| \bar{\sigma}_{\bar{yy}} - \tilde{\sigma}_{\bar{yy}} \right\| +\lambda_8 \left\| \bar{\sigma}_{\bar{xy}} - \tilde{\sigma}_{\bar{xy}} \right\|,  \label{eqs:SequentialTraining}
\end{split}
\end{equation}
as summarized in Algorithm 1. Although more robust compared to the simultaneous-solution form, the sequential strategy incurs in extended computational steps due to the operator-split iterations. 

% \begin{enumerate}
%     \item Generate random collocation set $\mathbf{X},\mathbf{T}$ for the whole spatial and temporal domain. 
%     \item if $itr=1$: ~~~~Initialize $\bar{\sigma}_v$ to zero. \\
%           else: ~~~~~~~~~~~~~Evaluate $\bar{\sigma}_v$ on the new collocation points. 
%     \item Optimize for $\mc{L}_f$ by passing $\bar{\sigma}_v$ as inputs to the optimizer. 
%     \item Evaluate $\bar{p}$ on the new collocation points. 
%     \item Optimize for $\mc{L}_s$ by passing $\bar{p}$ as input to the optimizer. 
%     \item Repeat until convergence. 
% \end{enumerate}

\begin{algorithm}\label{alg1}
\caption{Sequential fixed-stress-split algorithm for single-phase poroelasticity}\label{alg:cap}
\begin{algorithmic}[1]
\State $\mathbf{X}, \mathbf{T} \gets $ Sample uniformly spatial and temporal domains. 
\State $\bs{\theta}^n_{\bar{p}}, \bs{\theta}^n_{\bar{u}_{\bar{x}}},\bs{\theta}^n_{\bar{u}_{\bar{y}}},\bs{\theta}^n_{\bar{\varepsilon_v}} \gets $ Initialize randomly using Glorot scheme. 
\State $n \gets 1$
\State $\bar{\sigma}_v^0 \gets \bs{0}$ 
\While{$ \text{err} > \text{TOL}$}
    \State $\bs{\theta}^n_{\bar{p}} \gets$ Optimize $\mc{L}_f$ for $\bs{\theta}_{\bar{p}}$ over $\mathbf{X}, \mathbf{T}, \bar{\sigma}_v^{n-1}$.
    \State $\bar{p}^n \gets $ Evaluate $\bar{p}$ using $\bs{\theta}^n_{\bar{p}}$. 
    \State $\bs{\theta}^n_{\bar{u}_{\bar{x}}},\bs{\theta}^n_{\bar{u}_{\bar{y}}},\bs{\theta}^n_{\bar{\varepsilon_v}} \gets $ Optimize $\mc{L}_s$ for $\bs{\theta}_{\bar{u}_{\bar{x}}},\bs{\theta}_{\bar{u}_{\bar{y}}},\bs{\theta}_{\bar{\varepsilon_v}}$ over $\mathbf{X}, \mathbf{T}, \bar{p}^{n}$. 
    \State $\bar{\sigma}_v^n \gets $ Evaluate $\bar{\sigma}_v$ using $\bs{\theta}^n_{\bar{u}_{\bar{x}}},\bs{\theta}^n_{\bar{u}_{\bar{y}}},\bs{\theta}^n_{\bar{\varepsilon_v}}$. 
    \State $\bs{\theta}^n \gets \{\bs{\theta}^n_{\bar{p}}, \bs{\theta}^n_{\bar{u}_{\bar{x}}},\bs{\theta}^n_{\bar{u}_{\bar{y}}},\bs{\theta}^n_{\bar{\varepsilon_v}} \}$ as the collection of all parameters. 
    \State $\text{err} \gets \|\bs{\theta}^n - \bs{\theta}^{n-1}\| / \|\bs{\theta}^n\|$ with $\| \circ \|$ as the $L^2$ norm. 
    \State $n \gets n+1$
\EndWhile
\end{algorithmic}
\end{algorithm}

\subsection{PINN solution of two-phase poroelasticity}
For the case of two-phase poroelasticity, the main unknown variables are $u_x, u_y, p_c, p_w$. For the same reason explained earlier, we also introduce a separate unknown for the $\varepsilon_v$. Hence, the neural networks for the dimensionless form of these variables are
\begin{align}
    \bar{u}_{\bar{x}} &: (\bar{x}, \bar{y}, \bar{t}) \mapsto \mathcal{N}_{\bar{u}_{\bar{x}}}(\bar{x}, \bar{y}, \bar{t}; \bs{\theta}_{\bar{u}_{\bar{x}}}), \\
    \bar{u}_{\bar{y}} &: (\bar{x}, \bar{y}, \bar{t}) \mapsto \mathcal{N}_{\bar{u}_{\bar{y}}}(\bar{x}, \bar{y}, \bar{t}; \bs{\theta}_{\bar{u}_{\bar{y}}}), \\
    \bar{p}_c &: (\bar{x}, \bar{y}, \bar{t}) \mapsto \mathcal{N}_{\bar{p}_c}(\bar{x}, \bar{y}, \bar{t}; \bs{\theta}_{\bar{p}_c}), \\
    \bar{p}_w &: (\bar{x}, \bar{y}, \bar{t}) \mapsto \mathcal{N}_{\bar{p}_w}(\bar{x}, \bar{y}, \bar{t}; \bs{\theta}_{\bar{p}_w}), \\
    \bar{\varepsilon}_v &: (\bar{x}, \bar{y}, \bar{t}) \mapsto \mathcal{N}_{\bar{\varepsilon}_v}(\bar{x}, \bar{y}, \bar{t}; \bs{\theta}_{\bar{\varepsilon}_v}). 
\end{align}
The total loss function can be derived by following the single-phase case. We also follow two strategies here to solve the optimization problem. In the first case, we perform the optimization on the scored-adaptive-weights introduced earlier. The alternative approach is to decouple the displacement relations from the fluid relations, which results into two optimization problems that are solved sequentially following the stress-split strategy. Since the overall framework is very similar to the single-phase case, we avoid re-writing those details here.

\section{Applications}

In this section, we study the PINN solution of a few synthetic poromechanics problems. In particular, we first study Mandel's consolidation problem in detail to arrive the the right set of hyper-parameters for the PINN solution. We also discuss the sensitivity of the PINN solution to the choice of parameters for this problem. Following that, we apply PINN to solve Barry-Mercer's problem, and a synthetic two-phase drainage problem. In all of these cases, we keep the formulation in its two dimensional form, while considering coupled displacement and pressure effects. All the problems discussed below are coded using the SciANN package \cite{haghighat2021sciann}, and the codes will be shared publicly at \href{https://github.com/sciann/sciann-applications}{https://github.com/sciann/sciann-applications}.

\subsection{Mandel's problem}
\begin{figure}[H]
    \centering
    \includegraphics[width=0.8\textwidth]{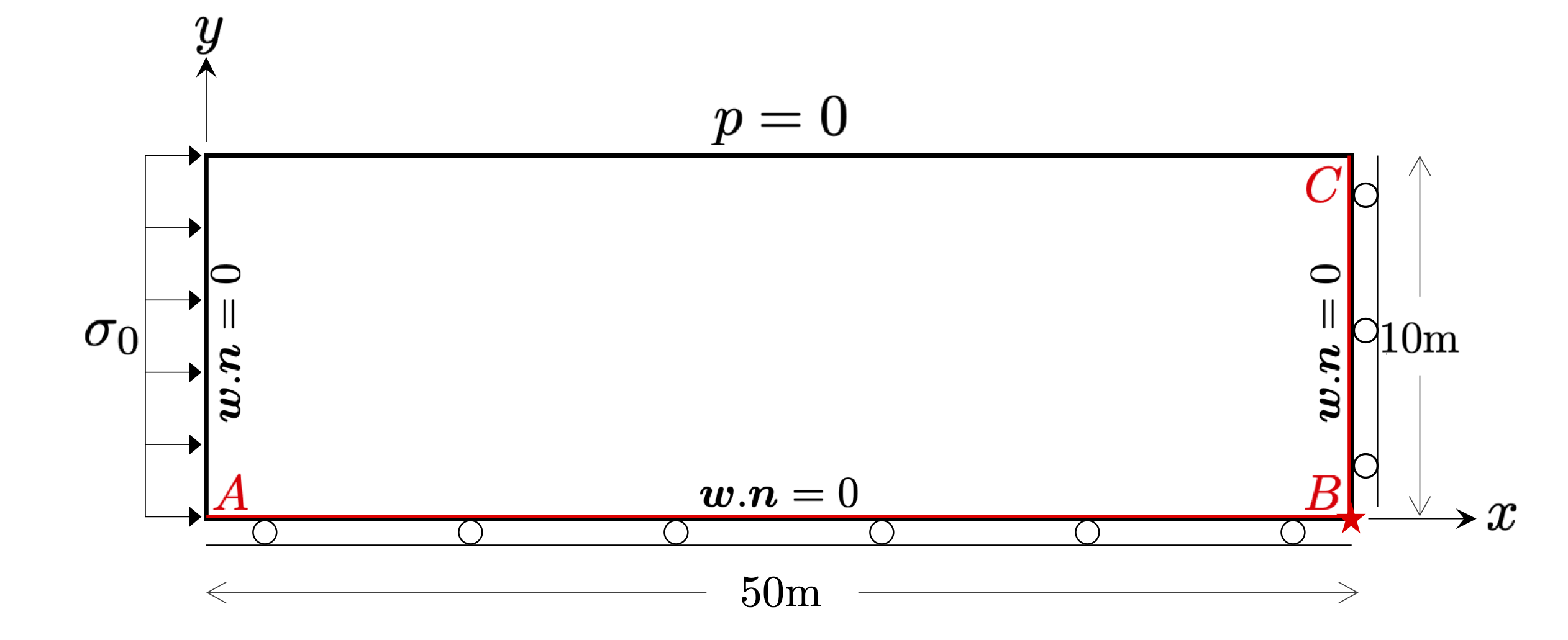}
    \caption{Mandel's consolidation problem. The left edge is subjected to $\sigma_0$ stress while to the top edge is kept traction free. Bottom and right edges are fixed in the normal direction. No-flux boundary condition is imposed on all edges except the top one, where a zero-pressure condition is imposed. }
    \label{fig:mandel-setup}
\end{figure}

We consider Mandel's problem as a reference benchmark in poroelasticity, as set up by \citet{jha2014coupled} (\cref{fig:mandel-setup}). The parameters of the problem are $\sigma_0 = -2~\text{MPa}$, $\rho_b=2400~\text{kg}/\text{m}^3$, $\rho_f=1000~\text{kg}/\text{m}^3$, ${\mu={0.6}\times {10^{-3}}\space \text{Pa.s}}$, $k=10^{-12}~\text{m}^2$, $E=120~\text{MPa}$, $\nu=0.25$, $b=1$, and $M=1200~\text{MPa}$. The dimensionless parameters are then evaluated as previously defined. In particular, we pay special attention to the coupling parameter $D^*$, which is $D^*=0.938$ for these parameters. Changing the Biot bulk modulus $M$ affects $D^*$. Due to the strong displacement and pressure coupling, the pressure rises above the applied stress and then dissipates over time. This problem allows us to examine the PINN method carefully under different considerations. The domain is subjected to constant stress $\sigma_0$ on the left edge, stress-free on the top edge, and fixed in normal-displacement on the bottom and right edges. All faces are subjected to no-flux condition except the top face which is subjected to $p=0$. The analytical solution for this problem is given in the \cref{sec:append1}.  

\begin{figure}[H]
    \centering
    \includegraphics[width=1\textwidth]{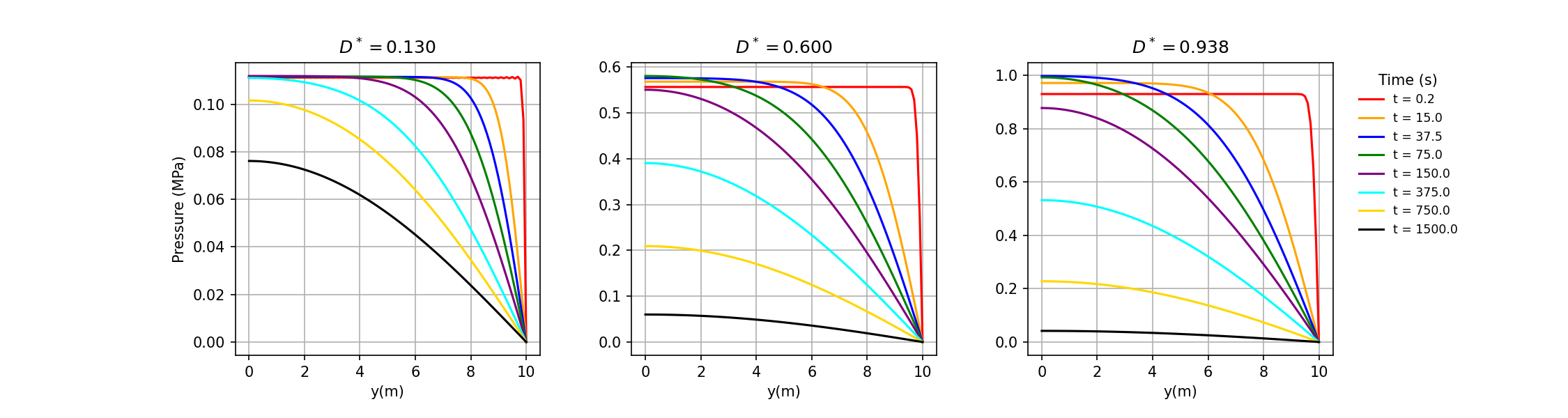}
    \includegraphics[width=1\textwidth]{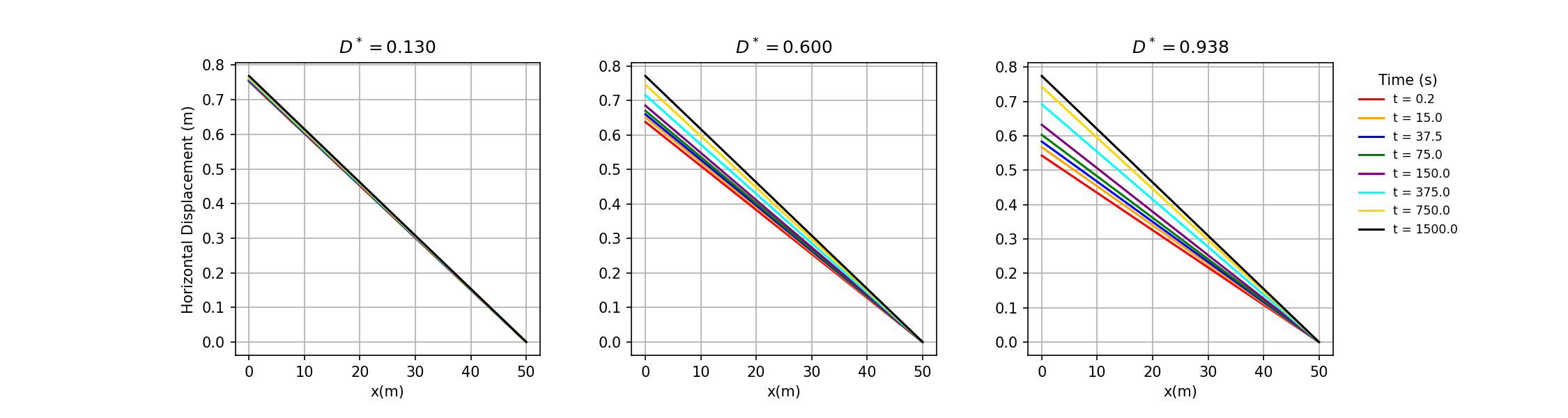}
    \includegraphics[width=1\textwidth]{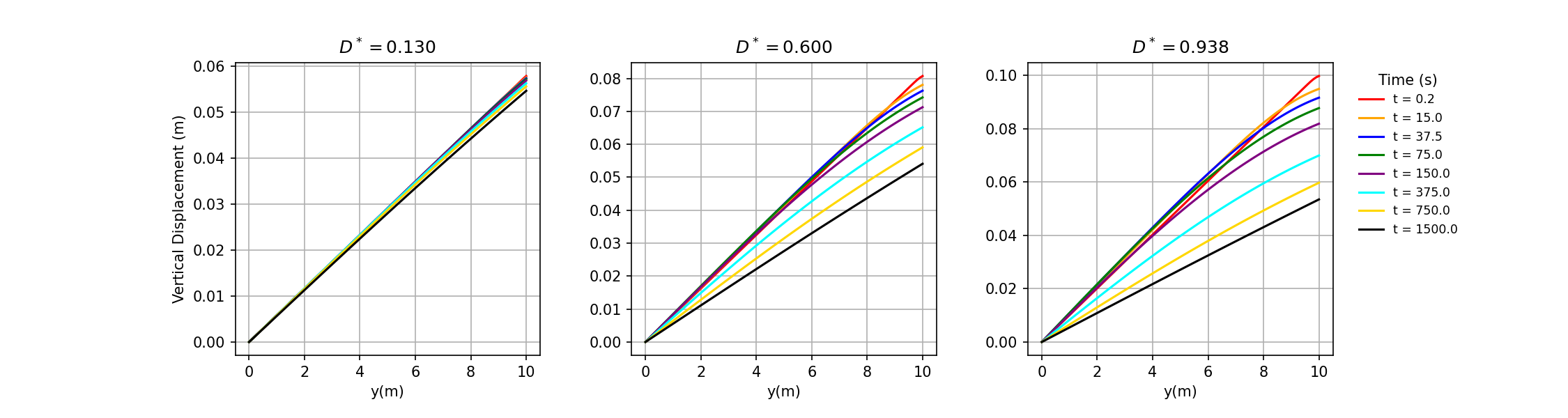}
    \caption{Pore pressure (top), horizontal displacement (middle), and vertical displacement (bottom) for Mandel's consolidation problem at different times and for different values of $D^*$ (left to right, $D^*=0.13,~0.6,~0.938$). Pressure and vertical displacement plots are evaluated along the BC line, while horizontal displacement is evaluated along the AB line, as shown in \cref{fig:mandel-setup}.}
    \label{fig:mandel_fig2}
\end{figure}

\begin{figure}[H]
    \centering
    \includegraphics[width=1.\textwidth]{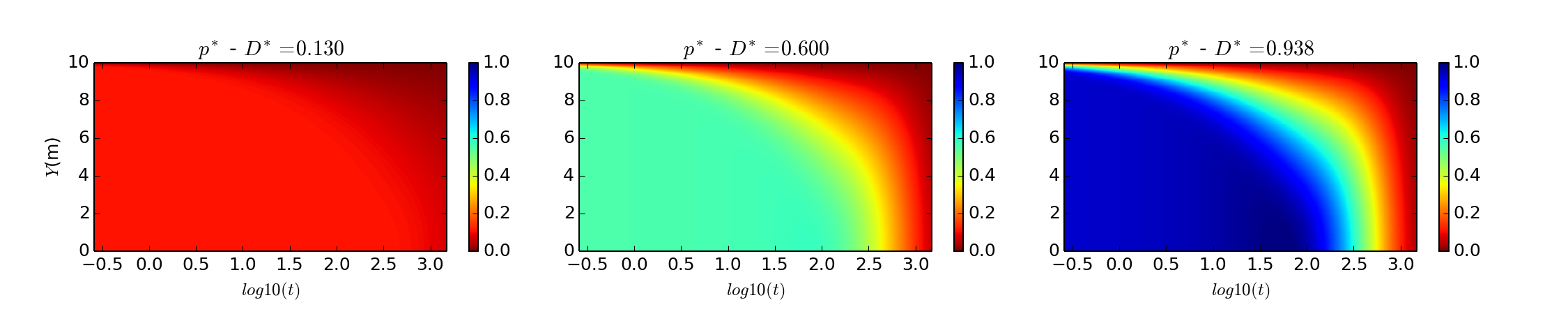}
    \includegraphics[width=1.\textwidth]{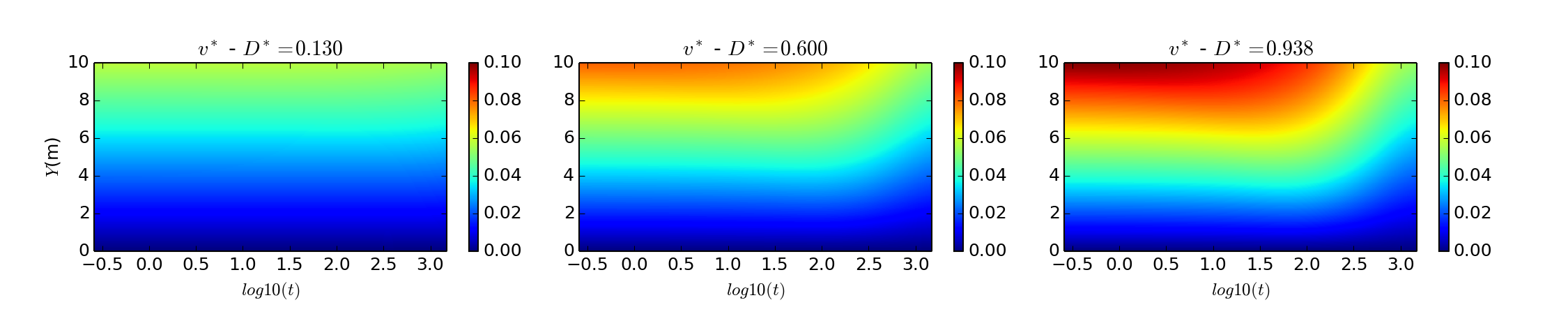}
    \caption{$y-\log t$ contour plots of pressure (top) and vertical displacement (bottom) along BC line (Y-axis)  for different values of $D^*$ for the Mandel's consolidation problem. }
    \label{fig:mandel_fig3}
\end{figure}

The analytical solution to Mandel's problem for different value of $D^*$ is plotted in \cref{fig:mandel_fig2}. Depending on the choice of $M$ (which changes $D^*$), the pressure may exhibit a strong response, almost of the same order as the applied overburden stress, or a weak response to the applied stress. If the coupling is strong (high $D^*$), the displacement field also shows a time-varying response. The vertical displacement exceeds the equilibrium displacement due to the change in pore-pressure and gradually returns back to its equilibrium state. These effects are also clear in the $y-log(t)$ plots, along the BC line, shown in \cref{fig:mandel_fig3}.

\subsubsection{Training strategies: simultaneous versus stress-split sequential and the role of adaptive weights}
\begin{figure}[H]
    \centering
    \includegraphics[width=\textwidth]{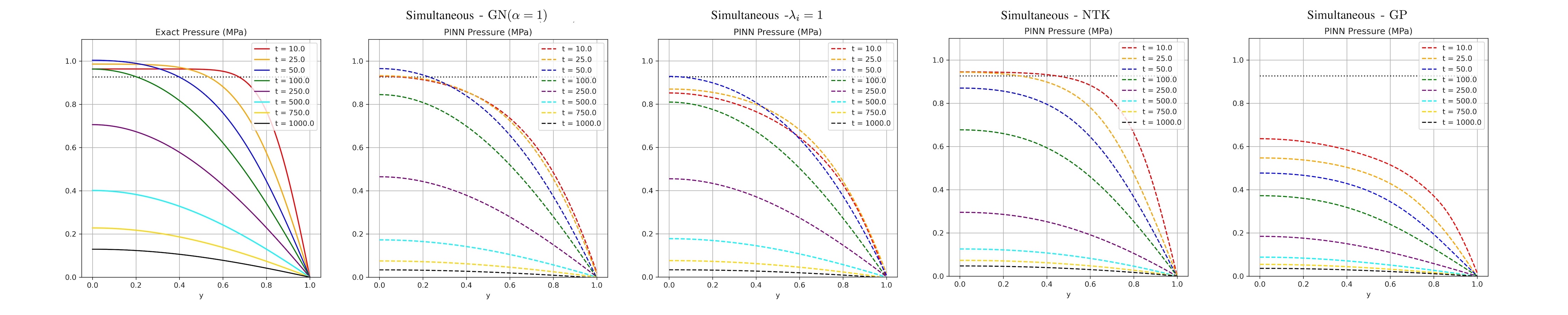}
    \includegraphics[width=\textwidth]{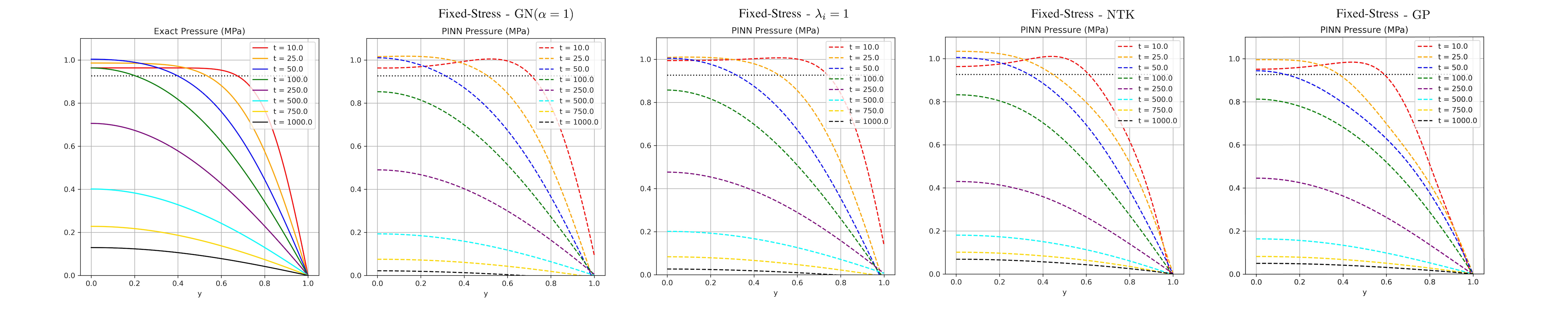}
    \caption{PINN solution for simultaneous (top) and fixed-stress-split sequential (bottom) training, considering adaptive weight schemes including GradNorm ($\alpha=1$), uniform weights with $\lambda_i=1$, Neural Tangent Kernel (NTK), and Gradient Pathologies (GP).  }
    \label{fig:mandel-coupled-weights}
\end{figure}

We study the convergence of the PINN solution using two different strategies for training the neural network: (1) Simultaneous training using the total loss function (\cref{eqs:SimultaneousTraining}); and (2) Sequential training using the fixed-stress split loss function (\cref{eqs:SequentialTraining}). We consider unweighted optimization, i.e., $\lambda_i=1$, non-scored adaptive weights based on gradient scaling (referred to as GP) \cite{wang2020understanding} and based on neural tangent kernel (referred to as NTK) \cite{wang2020ntk}, and the scored-adaptive-weights GradNorm algorithm \cite{chen2018gradnorm} discussed above (referred to as GN). We report only the evolution of pore pressure as a function of time along the vertical BC line as shown in \cref{fig:mandel-setup}. The results of the simultaneous and sequential trainings are summarized in \cref{fig:mandel-coupled-weights}. The most important consideration is if the method can capture the overpressure caused by the Mandel's setup. We observe that the GradNorm approach with $\alpha=1$ consistently yields the best results, while keep other optimization hyper-parameters unchanged. We additionally find that the fixed-stress sequential approach results in better performance compared to other cases. Therefore the training strategy we adopt for the rest of the paper is based on the use of GradNorm weights with $\alpha=1$ and using the sequential fixed-stress-split strategy.  

\subsubsection{Sequential solution: fixed-stress-split vs. fixed-strain-split}
\begin{figure}[H]
    \centering
    \includegraphics[width=\textwidth]{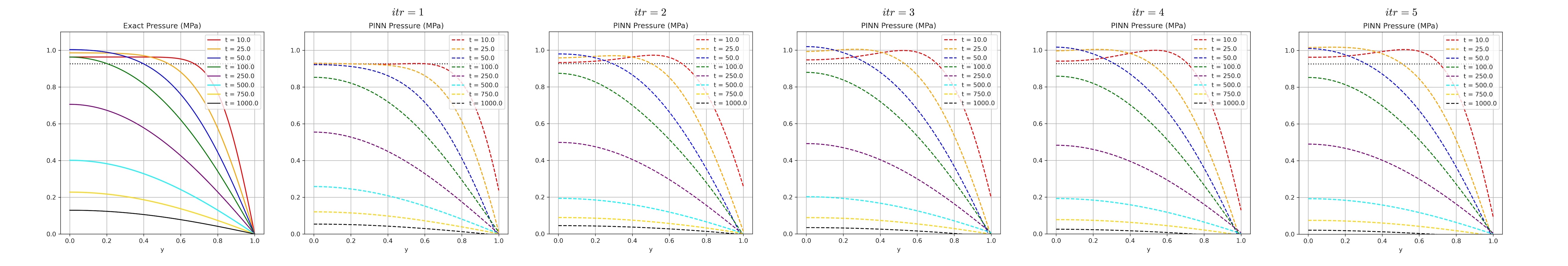}
    \includegraphics[width=\textwidth]{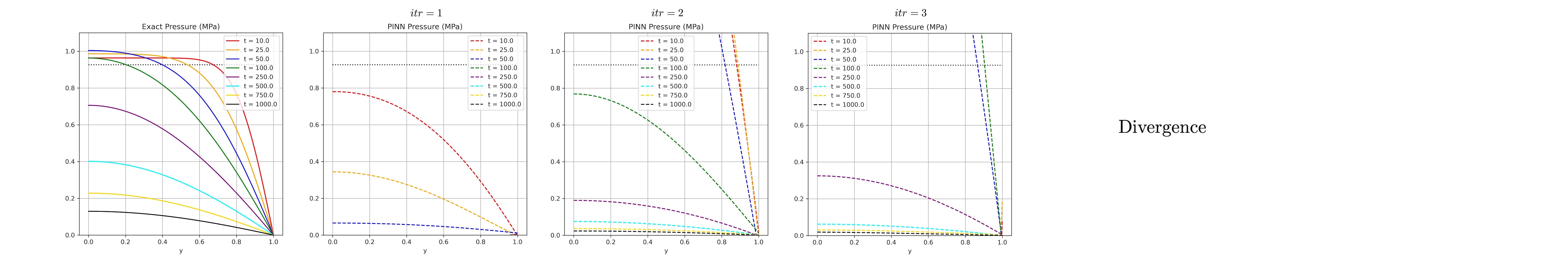}
    \caption{PINN solution for Mandel's problem using sequential stress-split (top) versus strain-split (bottom) network training. The strain-split approach is not stable for $D^*=0.934$ and results in divergence.  }
    \label{fig:mandel_fig4}
\end{figure}

Here, we explore the role of the fixed-stress-split and fixed-strain-split strategies for the sequential solution of the coupled system. The two equations describing the flow problem are almost identical, the key difference being the variable passing the information related to the mechanics problem. The results are shown in \cref{fig:mandel_fig4}. The top plots show the stress-split iterations while the bottom plots depict the strain-split iterations. We note that the strain-split strategy diverges while the stress-split formulation converges to the right solution. We note that if we choose $D^*$ based on the stability criterion reported by \citet{kim2011stability}, the strain-split strategy also converges to the expected solution. {This is a surprising finding of our study because the stability criterion was originally developed for Euler integration of the FEM discretized system while the approximation space of PINNs remains continuous, both in space and time.}

\subsubsection{PINN's $D^*$-dependence}
\begin{figure}[H]
    \centering
    \includegraphics[width=0.7\textwidth]{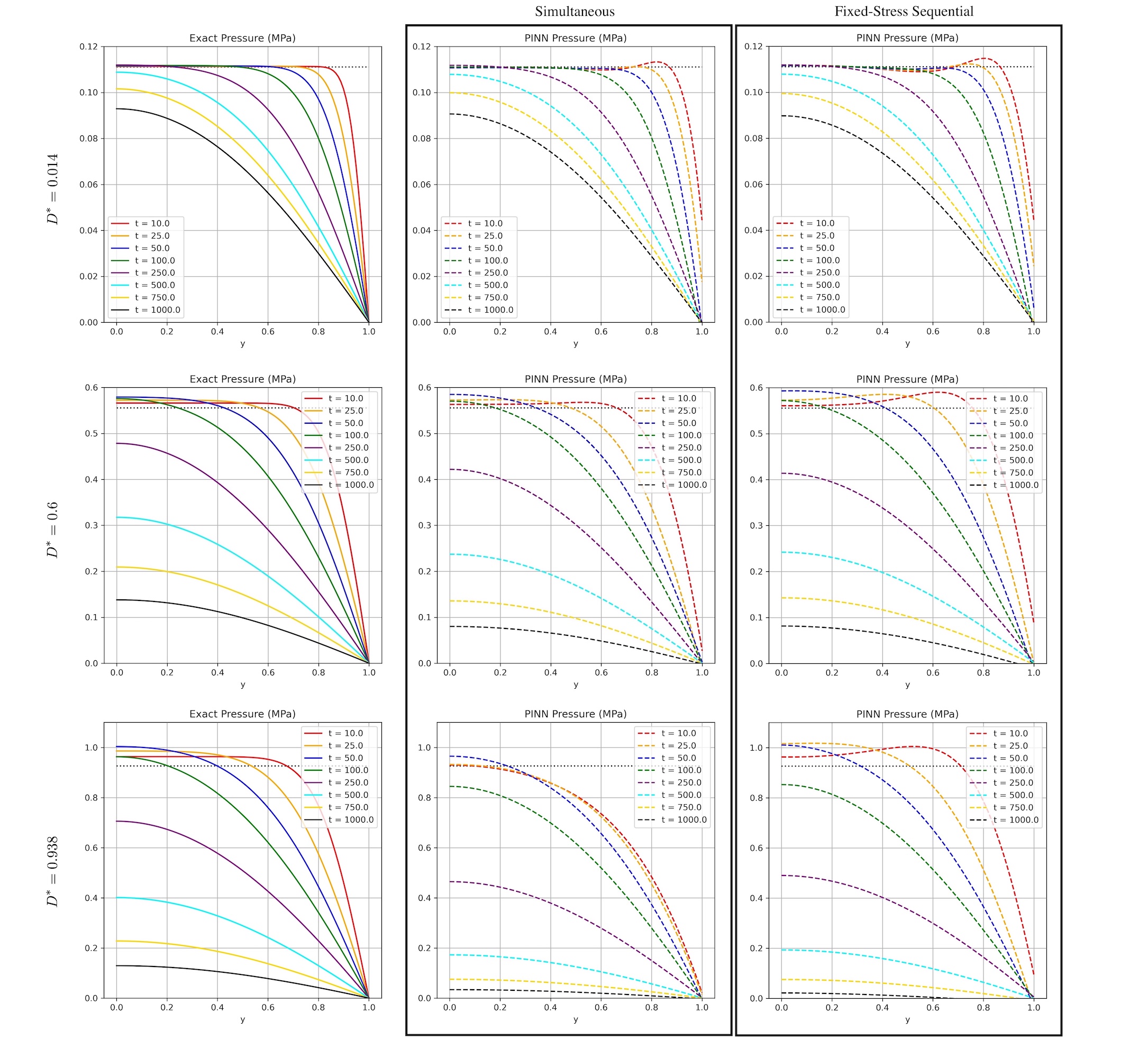}
    \caption{PINN solution for Mandel's problem and its dependence of $D^*$ for simultaneous (middle) and fixed-stress-split sequential (right) solution strategies. }
    \label{fig:mandel_fig5}
\end{figure}

As it is clear from \cref{eqs:nondim4}, $D^*$ controls the degree of coupling between flow and mechanics problems, with values ranging in $D^* \in (0, 1)$. A small value of $D^*$ implies no coupling, while a large value implies strong coupling. Here, we assess the quality of the PINN solution as a function of $D^*$. The results are plotted in \cref{fig:mandel_fig5}. Not surprisingly, the PINN solutions have highest accuracy for small $D^*$. For a fixed set of hyper parameters, the accuracy gradually decreases as $D^*$ increases. Note that all of trainings are conducted using the same number of epochs. An equivalent interpretation is that as $D^*$ increases, the training requires increasingly more epochs to reach the same accuracy. This observation confirms the challenges associated with PINN solvers for coupled multiphysics problems. 

\subsubsection{Network size and architecture}
It is no surprise that the network size and architecture, as well as the optimizer's hyper parameters such as learning rate, can also play a critical role in the quality of PINN solutions. For all the examples shown in this manuscript, we have set up neural networks with 4 hidden layers and with 100 neurons in each layer, using the hyperbolic-tangent activation function. The optimizer is Adam, with an initial learning rate of $10^{-3}$, and with an exponential learning decay to $10^{-5}$ at the end of training. We have also experimented with the Fourier feature architecture \cite{10-PINN-FourierNetwork}, ResNet architecture, globally and locally adaptive activation functions \cite{2-PINN-LocallyAdaptiveActivationFunction}, and SIREN architecture with periodic activation functions \cite{sitzmann2020implicit}. However, we have not seen any significant improvement for solving the coupled flow-mechanics problem studied here.

%\newpage
\subsection{Barry-Mercer's Injection-Production Problem}
\begin{figure}[H]
    \centering
    \includegraphics[width=.6\textwidth]{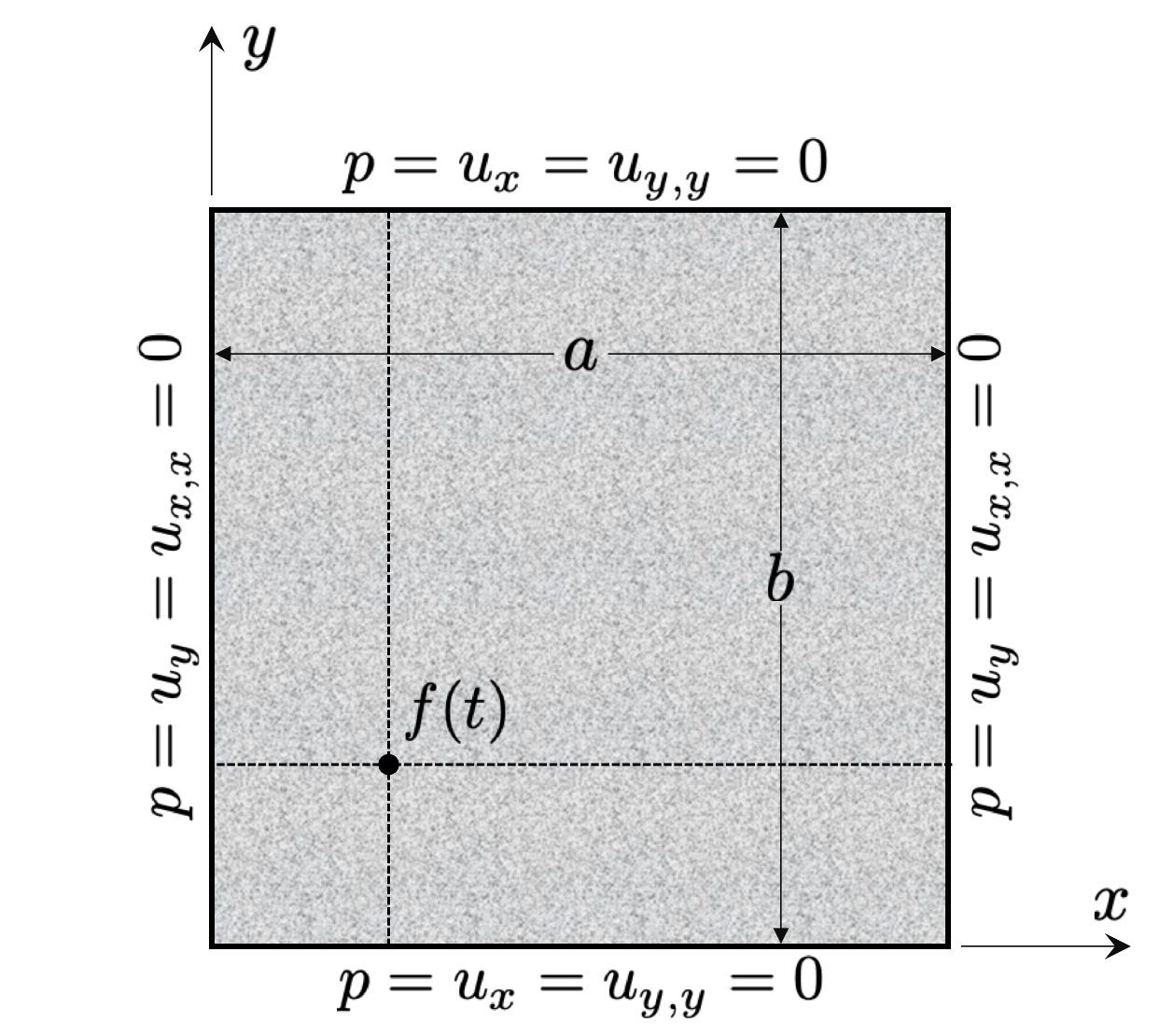}
    \caption{Barry-Mercer's injection-production problem. All edges are subjected to zero-pressure condition, and they are fixed in their normal direction. A production-injection source term $f(t)$ is applied at $(x_0, y_0)$.  }
    \label{figs:barry_mercer_1}
\end{figure}

Practical problems of porous media involve injection and production of fluids. A classical benchmark problem of this type is the Barry-Mercer problem \cite{Barry-Mercer}, in which a time-dependent fluid injection/production well is considered. In this two-dimensional problem, the gravity term is ignored. The solid and fluid phases are considered incompressible, and Biot’s coefficient is assumed ${b=1}$, which results in ${D^*=1.0}$. The initial conditions are zero pressure and displacement in the entire domain. The boundary conditions are depicted in \cref{figs:barry_mercer_1}. The analytical solution to this problem is given in the \cref{sec:append2}. 

In this study, similar to \cite{phillips2005finite}, the location of the injection/production point is set equal to ${(x_0,y_0) = (0.25,0.25)}$, and the domain length and width are assumed ${a=b=1\text{m}}$. Elastic properties of the solid phase are taken as ${E=4.67 ~\text{MPa}}$ and ${\nu=0.167}$. The permeability and the fluid viscosity are ${k=10^{-10}~ \text{m}^2}$ and ${\mu=10^{-3}~ \text{Pa.s}}$. The injection/production function $f(t)$ is given by:
\begin{align}
f(t) = {2 \beta}{\delta(x-x_0)\delta(y-y_0)}{\sin(\beta t)},~~~~~\text{where}~~~~~ {\beta} = \frac{{E(1-\nu)}k}{(1+\nu)(1-2\nu)ab\mu},
\end{align}
in which ${\delta}$ denotes the Dirac delta function and its Gaussian approximation is used here, i.e., ${\delta(x)}=\frac{1}{\alpha \sqrt{\pi}}{e^{-(x/\alpha)^2}}$. In this study, we assume ${\alpha=0.04}$. Note that higher values of $\alpha$ diffuse the injection/production source while smaller values concentrate the source function and make the training increasingly challenging. The temporal domain is taken as ${\hat{t} \in [0,2\pi]}$, where ${\hat{t}}$ is the dimensionless time defined as ${\hat{t} = \beta t}$ \cite{Barry-Mercer}. 

\begin{figure}[H]
    \includegraphics[width=1\textwidth]{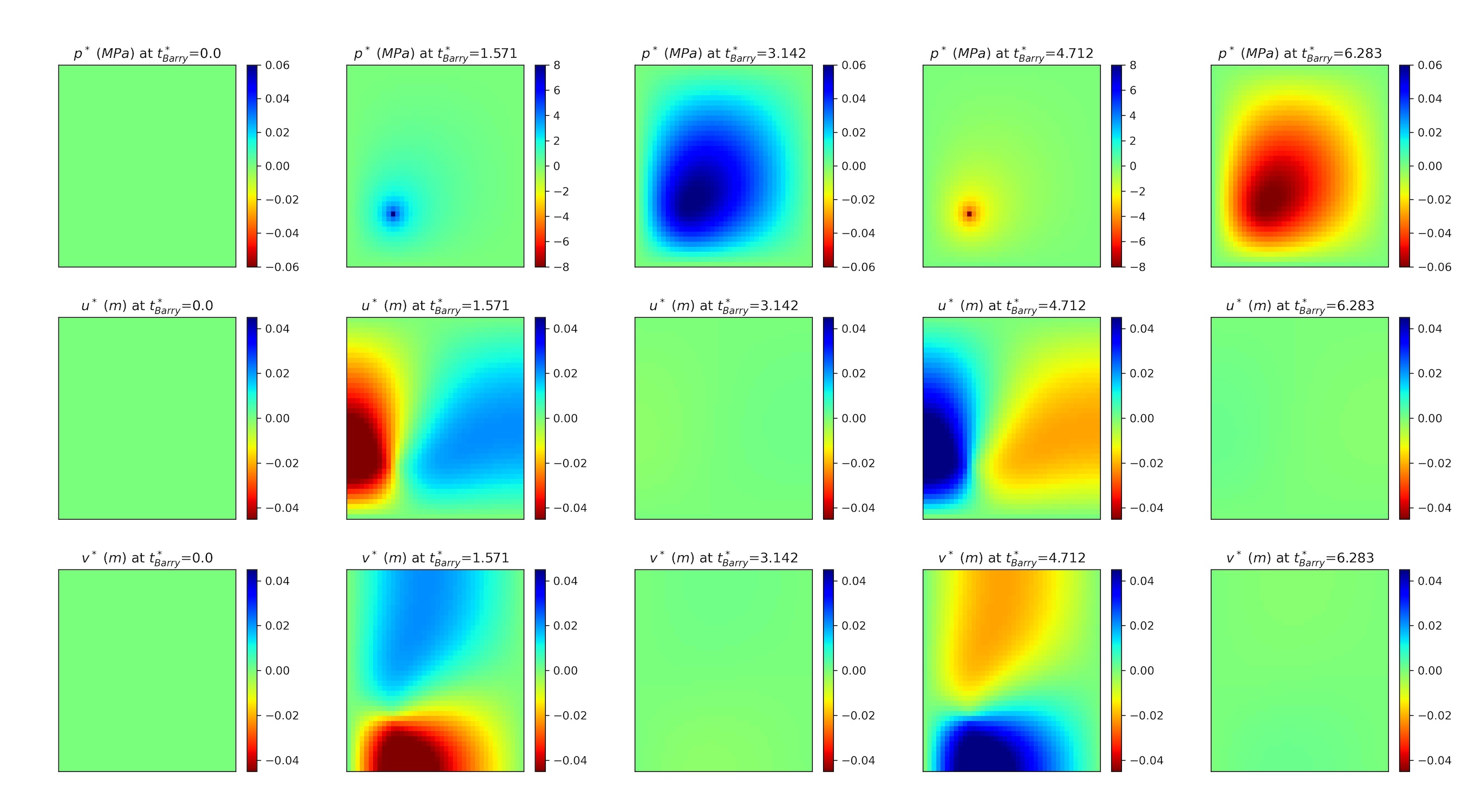}
    \caption{Barry-Mercer's analytical solution for pore pressure (top), horizontal (middle) and vertical (bottom) displacements, respectively. Each column represents a different time.  }
    \label{figs:barry_mercer_2}
\end{figure}

\begin{figure}[H]
    \includegraphics[width=1\textwidth]{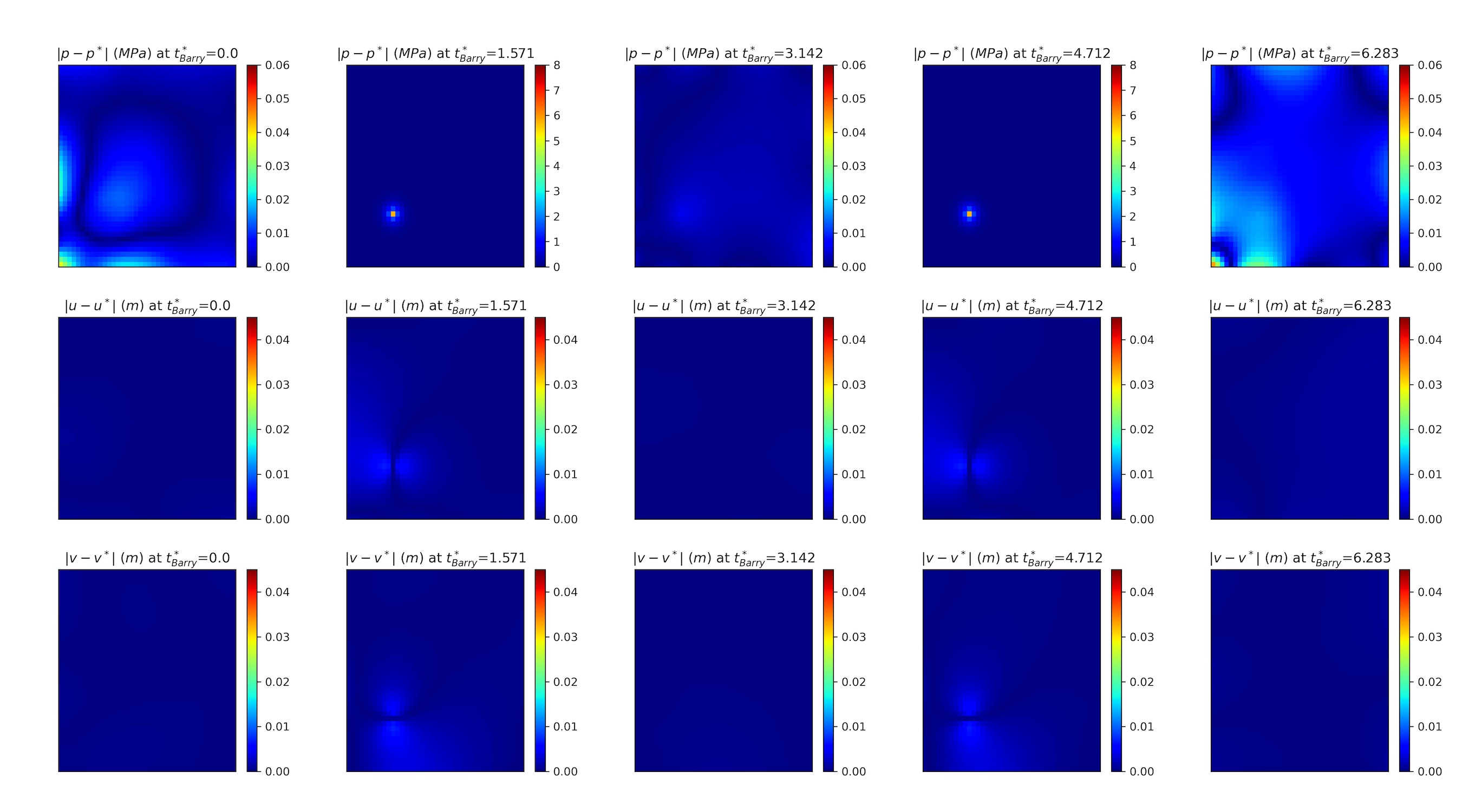}
    \caption{Absolute error plots of Barry-Mercer's PINN solution for pore pressure (top), horizontal (middle) and vertical (bottom) displacements, respectively, obtained using the fixed-stress-split training strategy. Each column represents a different time. Note that the left two columns represent the solution values at the boundaries. }
    \label{figs:barry_mercer_3}
\end{figure}

\begin{figure}[H]
    \includegraphics[width=1\textwidth]{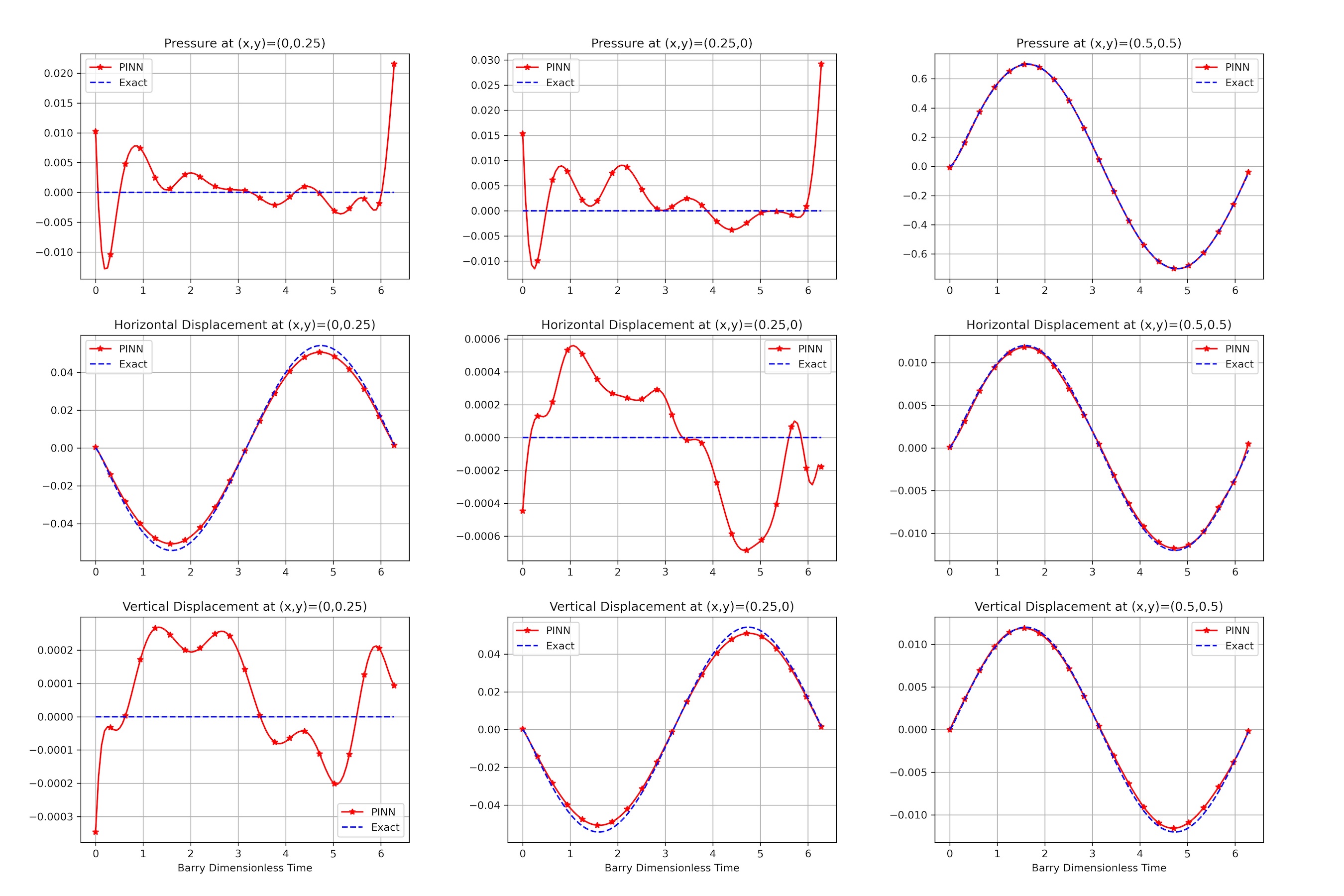}
    \caption{Barry-Mercer's PINN solution for pore pressure (top), horizontal displacement (middle), and vertical displacement (bottom), respectively, obtained using the fixed-stress-split training strategy. Each column represents a different location around the injection-production well. }
    \label{figs:barry_mercer_4}
\end{figure}

The analytical solution is plotted in \cref{figs:barry_mercer_2}. The absolute error plots of the stress-split sequential training are plotted in \cref{figs:barry_mercer_3} and their line plots at 3-points are plotted in \cref{figs:barry_mercer_4}. As you can see, compared to the results reported in earlier studies, PINN solution has a great agreement with the expected solution. 

% \begin{figure}[H]
%     \includegraphics[width=1\textwidth]{figs/barrymercer/Sequential_4x100LrExpF40000S40000GNalpha1_PUV_itr08.jpg}
%     \caption{Barry-Mercer's PINN solution for pore pressure (top), horizontal (middle) and vertical (bottom) displacements, respectively. Each column represents a different time. Note that the left two columns represent the solution values at the boundaries. }
%     \label{figs:barry_mercer_3}
% \end{figure}

%\newpage
\subsection{Two-Phase Drainage Problem}
\begin{figure}[H]
    \centering
    \includegraphics[width=1.\textwidth]{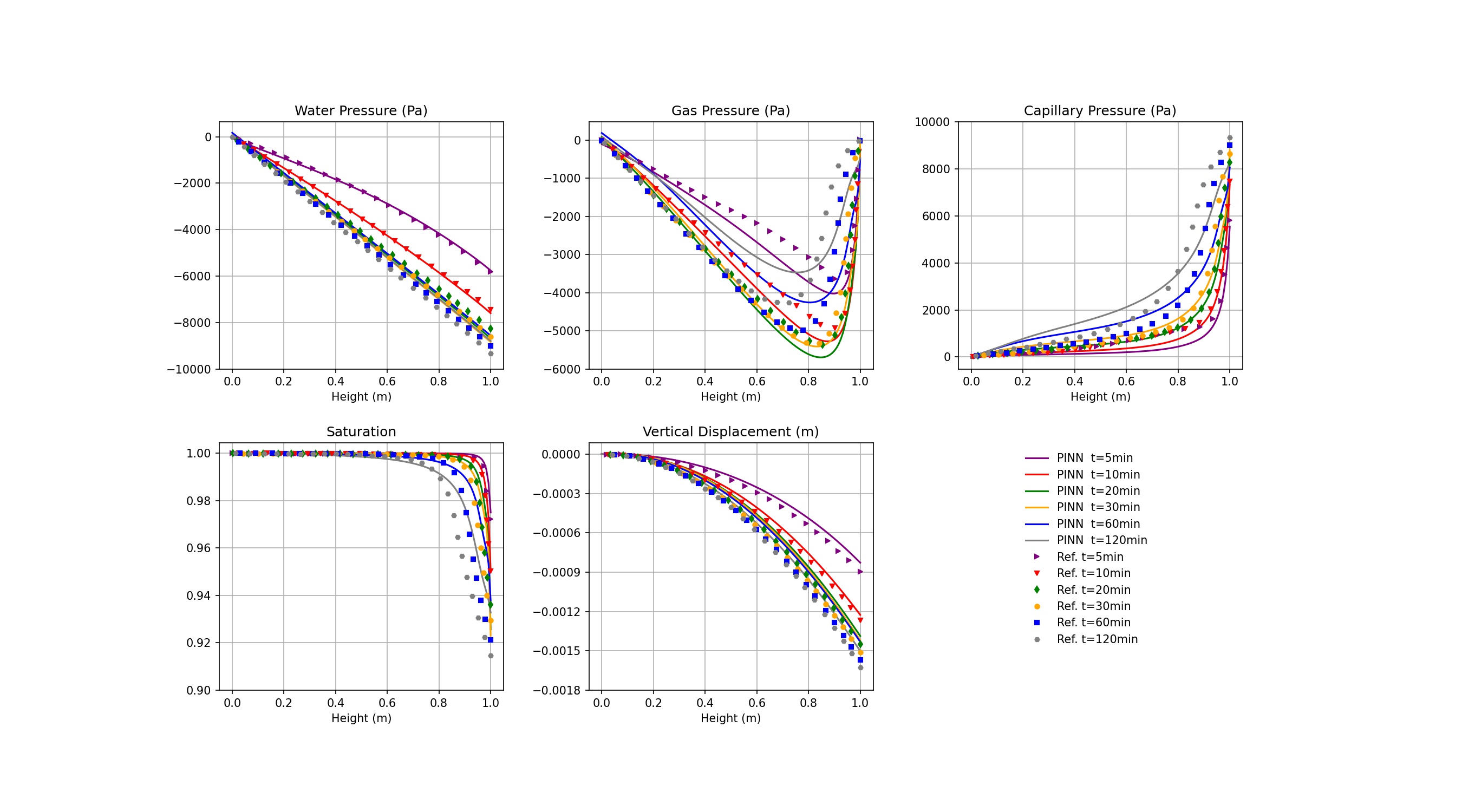}
    \caption{The PINN solution for the two-phase soil column water drainage problem, at different times. The reference solution is reproduced form the results reported by \citet{lewis1998finite}. }
    \label{figs:twophase_1}
\end{figure}

As a third example, we consider a synthetic two-phase flow problem, in which water drains from the bottom of a soil column due to gravity. This well-known example was experimentally investigated by \citet{liakopoulos1964transient} and has been used as a benchmark problem \cite{lewis1998finite}. To impose one-dimensional boundary conditions in the laboratory test, the soil column was placed in an impermeable vessel open at the bottom and top. The column is initially fully saturated with water and in mechanical equilibrium. Then the water inflow is ceased and a no-displacement boundary condition at the bottom together with the free-stress condition at the top surface is imposed. The water pressure at the bottom and gas pressure at both the bottom and top surfaces are set equal to atmospheric pressure. It is assumed that this ${1\space\text{m}}$ soil column has linear elastic behavior, in which the Young modulus is ${E=1.3 \space\text{MPa}}$, the Poisson ratio is ${\nu=0.4}$, the porosity is ${\phi=0.2975}$, and the Biot coefficient is ${b=1}$. The densities of solid, water, and gas phases are ${\rho_s=2400 \space \text{kg}/\text{m}^3}$, ${\rho_w=1000 \space \text{kg}/\text{m}^3}$, and ${\rho_g=1.2 \space \text{kg}/\text{m}^3}$, respectively. The fluid compressibilities  are ${c_w={{5}\times 10^{-10}}\space \text{Pa}^{-1}}$, ${c_g= {10^{-5}}\space \text{Pa}^{-1}}$, with the solid bulk modulus ${K_s= {10^{12}}\space \text{Pa}}$. In this numerical test, the intrinsic permeability is ${k={4.5}\times {10^{-13}}\space\text{m}^2}$, the gravitational acceleration is ${g={9.806}\space\text{m}/\text{s}^2}$, and the atmospheric pressure is zero. Furthermore, although the saturation relation, as well as relative permeability function for the water phase, are obtained from experimental expressions, the relative permeability of the gas phase is calculated by Brooks-Corey \cite{Brooks-Corey} model as follows
\begin{align*}
%   {S_w}&=1-0.10152{(p_c/9806)^{2.4279}}, \\
%   {p_c}&=\rho_w g \left(\frac{1-S_w}{0.10152}\right)^\frac{1}{2.4279}, \\
  {p_c}&= 2.57 \rho_w g (1-S_w)^\frac{1}{2.4279}, \\
  {k_{rw}}&={1 - 2.2{(1-S_w)}}, \\
  {k_{rg}}&=\text{max}{[(1 - S_e)^{2}(1-S_e^{{(2+\lambda)}/{\lambda}}), 0.0001]}, \\
  {{S_e}}&={(S_w - S_{rw})}/{(1-S_{rw})}.
\end{align*}
where ${\lambda=3}$, denotes the pore size distribution, ${S_e}$ is the effective saturation, and ${S_{rw}=0.2}$ is the connate water saturation. In addition, the fluid viscosities are ${\mu_w={10^{-3}}\space \text{Pa.s}}$, ${\mu_g={1.8}\times {10^{-5}}\space \text{Pa.s}}$. The results are plotted in \cref{figs:twophase_1}, which are in agreement with the expected results. 

\section{Discussions and Concluding Remarks}

We study an application of Physics-Informed Neural Networks (PINNs) for the forward solution of coupled fully- and partially-saturated porous media. Our study is the first to consider fully coupled relations of porous media under single-phase and two-phase flow conditions. We report a dimensionless form of these relations that results in a stable and convergent behavior of the optimizer. 
% We also find here that the stress-split sequential training is the most suitable approach for solving porous media relations using PINNs. 
One of the key contributions from this work is proposing a novel sequential training of the neural network. In the current application to poromechanics problems, it takes the form of a fixed-stress split of the loss function. This training strategy results in enhanced convergence and robustness of the PINN formulation.
% Additionally, after a careful consideration of various adaptive weighting strategies, we report a  suitable algorithm for solving such problems. 

While our results show good agreement with expected results, we still find that training PINNs is very slow and control over its accuracy is challenging. 
% We find that with more terms in the total loss function, i.e., considering fully-coupled relations, training PINNs become an impossible task. Therefore, 
As reported by others, we associate the training challenge to the multi-objective optimization problem and the use of a first-order optimization method. Considering the challenges faced and resolved in this study for the forward problem, a next step is to apply PINNs for inverse problems of coupled multiphase flow and poroelasticity.

% \section*{Acknowledgements}

\appendix
\section{Analytical solutions}
{For completeness, we report the analytical solutions that exist for Mandel's problem as well as for Barry-Mercer's problem. A summary of these solutions are also discussed in} \cite{castelletto2015accuracy, phillips2005finite}.

\subsection{Mandel's analytical solution}\label{sec:append1}

The analytical solution to the Mandel's consolidation problem is expressed as 

\begin{align*}
{p(x,y,t)} &= \frac{2|\sigma_0| B (1+\nu_u)}{3} \left( \sum_{i=1}^{\infty} \frac{\sin \alpha_i}{\alpha_i - \sin\alpha_i\cos\alpha_i} ~ ({\cos (\frac{\alpha_i y}{L_y})-{\cos\alpha_i})}~ \exp[\frac{-\alpha_i^2 c_f t}{L_y^2}]\right), \\
{v(x,y,t)} &= \frac{\sigma_0 y}{G} \left(\frac{\nu}{2} - {\nu_u} \sum_{i=1}^{\infty} \frac{\sin\alpha_i\cos\alpha_i}{\alpha_i - \sin\alpha_i\cos\alpha_i} ~ \exp[\frac{-\alpha_i^2 c_f t}{L_y^2}]\right) \\
&+\frac{\sigma_0 L_y}{G} \left( \sum_{i=1}^{\infty} \frac{\cos \alpha_i}{\alpha_i - \sin\alpha_i\cos\alpha_i} ~ {\sin (\frac{\alpha_i y}{L_y})}~ \exp[\frac{-\alpha_i^2 c_f t}{L_y^2}]\right), \\
{u(x,y,t)} &=\frac{\sigma_0(L_x-x)}{G}\left(\frac{-(1-\nu)}{2} + (1-\nu_u) \sum_{i=1}^{\infty} \frac{\sin\alpha_i \cos\alpha_i}{\alpha_i - \sin\alpha_i\cos\alpha_i} ~ \exp[\frac{-\alpha_i^2 c_f t}{L_y^2}]\right).
\end{align*}
where the parameters are 
\begin{align*}
{B} &= \frac{bM}{K_{dr}+{b^2}{M}}, \\
{\nu_u} &= \frac{3\nu+bB(1-2\nu)}{3-bB(1-2\nu)}, \\
{c_f} &= \frac{k}{\mu(1/M+{b^2}/{K_{dr}})}, \\
{G} &= \frac{E}{2(1+\nu)}
\end{align*}
It must be noted that ${\sigma_0}$ is negative for compression loading and positive for tension loading, and ${\alpha_i}$ are the roots of the equation:
\begin{equation*}
{\tan \alpha_i}=\frac{1-\nu}{\nu_u - \nu}{\alpha_i}.
\end{equation*}

\subsection{Barry-Mercer's problem}\label{sec:append2}
Assuming the parameters
\begin{align*}
{\lambda_n}&={n \pi}, \\
{\lambda_q}&={q \pi}, \\
{\lambda_{nq}}&={\lambda_n^2}+{\lambda_q^2}, \\
{\beta} &= \frac{{E(1-\nu)}k}{(1+\nu)(1-2\nu)ab\mu}, 
\end{align*}
and with an injection function of the form
\begin{align*}
{Q} = {2 \beta}{\delta(x-x_0)\delta(y-y_0)}{\sin(\beta t)}
\end{align*}
with $(x_0,y_0)$ as source location. The time domain is considered as ${\hat{t}\in [0, 2\pi]}$, where ${\hat{t}=\beta t}$. Definition of the pressure, horizontal and vertical displacement in the transformed coordinate :
\begin{align*}
{\hat{p}(n,q,t)} &=\frac{-2 \sin(\lambda_n x_0) \sin(\lambda_q y_0)}{\lambda_{nq}^2 + 1}(\lambda_{nq}\sin(\beta t) - \cos(\beta t) + {e^{-\lambda_{nq} \beta t}}),\\
{\hat{u}(n,q,t)} &=\frac{\lambda_n}{\lambda_nq}{\hat{p}(n,q,t)}, \\
{\hat{v}(n,q,t)} &=\frac{\lambda_q}{\lambda_nq}{\hat{p}(n,q,t)},
\end{align*}
The real pressure, horizontal and vertical displacement in the real coordinates can be defined as:
\begin{align*}
{p(x,y,t)} &= \frac{-4E(1-\nu)}{(1+\nu)(1-2\nu)}{\sum_{n=1}^{\infty}}~{\sum_{q=1}^{\infty}{\hat{p}(n,q,t)} {\sin(\lambda_n x)} {\sin(\lambda_q y)}}, \\
{u(x,y,t)} &= {4}{\sum_{n=1}^{\infty}}~{\sum_{q=1}^{\infty}{\hat{u}(n,q,t)} {\cos(\lambda_n x)} {\sin(\lambda_q y)}}, \\
{v(x,y,t)} &= {4}{\sum_{n=1}^{\infty}}~{\sum_{q=1}^{\infty}{\hat{v}(n,q,t)} {\sin(\lambda_n x)} {\cos(\lambda_q y)}}.
\end{align*}

%% References
%%
%% Following citation commands can be used in the body text:
%% Usage of \cite is as follows:
%%   \cite{key}         ==>>  [#]
%%   \cite[chap. 2]{key} ==>> [#, chap. 2]
%%

%% References with bibTeX database:

\bibliographystyle{plainnat}

\bibliography{references}

\end{document}